\documentclass[5p,twocolumn,times]{elsarticle}

\makeatletter
\def\ps@pprintTitle{%
 \let\@oddhead\@empty
 \let\@evenhead\@empty
 \def\@oddfoot{}%
 \let\@evenfoot\@oddfoot}
\makeatother

\usepackage{hyperref}
\makeatletter
\g@addto@macro{\UrlBreaks}{\UrlOrds}
\makeatother

\usepackage{amsmath,amsfonts}

\usepackage{paralist}
\usepackage{booktabs} 
\usepackage{color}
\usepackage{bbm}
\usepackage{dsfont}
\usepackage{algorithm}
\usepackage[noend]{algpseudocode}
\usepackage{esvect}
\usepackage{multirow}
\usepackage{subcaption}
\usepackage{fontawesome}

\usepackage{epstopdf}
\usepackage{graphics}
\usepackage{graphicx}
\DeclareGraphicsExtensions{.eps,.gz,.eps,.pdf,.png,.jpg}
\graphicspath{{./}{./images/}}

\newcommand{\sstitle}[1]{\smallskip\noindent\textbf{#1.\/}}

\def\Snospace~{\S{}}

\makeatletter
\newcommand{\removelatexerror}{\let\@latex@error\@gobble}
\makeatother

\begin{document}

\setlength{\belowdisplayskip}{0pt}
\setlength{\belowdisplayshortskip}{0pt}
\setlength{\abovedisplayskip}{0pt}
\setlength{\abovedisplayshortskip}{0pt}

\begin{frontmatter}

\title{A War Beyond Deepfake: Benchmarking Facial Counterfeits and Countermeasures}

\author[hust]{Pham Minh Tam}
\author[griffith]{Huynh Thanh Trung}
\author[hust]{Tong Van Vinh}
\author[hannover]{Nguyen Thanh Tam}
\author[deakin]{Nguyen Thanh Thi}
\author[uq]{Hongzhi Yin}
\author[griffith]{Nguyen Quoc Viet Hung}

\address[hust]{Hanoi University of Science and Technology, Vietnam}
\address[griffith]{Griffith University, Australia}
\address[hannover]{Leibniz Universit{\"a}t Hannover, Germany}
\address[deakin]{Deakin University, Australia}
\address[uq]{The University of Queensland, Australia}

\begin{abstract} 

In recent years, visual forgery has reached a level of sophistication that humans cannot identify fraud, which poses a significant threat to information security. A wide range of malicious applications have emerged, such as fake news, defamation or blackmailing of celebrities, impersonation of politicians in political warfare, and the spreading of rumours to attract views. As a result, a rich body of visual forensic techniques has been proposed in an attempt to stop this dangerous trend. In this paper, we present a benchmark that provides in-depth insights into visual forgery and visual forensics, using a comprehensive and empirical approach. More specifically, we develop an independent framework that integrates state-of-the-arts counterfeit generators and detectors, and measure the performance of these techniques using various criteria. We also perform an exhaustive analysis of the benchmarking results, to determine the characteristics of the methods that serve as a comparative reference in this never-ending war between measures and countermeasures.  

\end{abstract}


\end{frontmatter}

\section{Introduction}

The development of fake visual content, such as fake images and fake videos, has undergone rapid proliferation in recent years~\cite{mirsky2021creation}. In particular, fake facial image generators have received particular interest, since the face plays an essential role in human communication and represents the identity of the person. Forged facial images and videos have reached such a high level of quality that even people with good vision, under ideal lighting conditions, cannot distinguish between fake and real items~\cite{rossler2019faceforensics++}. These facial manipulation techniques have been used in various malicious applications, such as counterfeit news generation, click-baits, impersonation and fraudulent transactions~\cite{li2020celeb,bi2018fast}.  

Given the importance of fake content detection, visual forensics has emerged, and can be categorised into two paradigms. On the one hand, \emph{computer vision} approaches rely on handcrafted features to detect anomalous patterns in visual content, including frequency-based techniques, visual artefacts, and techniques that examine head pose~\cite{durall2019unmasking,matern2019exploiting,yang2019exposing}. On the other hand, \emph{deep learning} approaches leverage the advances in relation to deep neural networks (DNNs) to automatically extract hidden features that go beyond human perception to distinguish real from fake visual contents. Techniques such as \textit{Mesonet}~\cite{afchar2018mesonet}, \textit{Capsule}~\cite{nguyen2019capsule}, \textit{Xception}~\cite{chollet2017xception} and \textit{GAN-fingerprint}~\cite{yu2019attributing} fall into this category.

However, recent advances in modern artificial intelligence have given rise to a new and ever-evolving class of visual forgery techniques. These techniques exploit the power of AI to hide the digital footprints generated by the forgery process, and can trick even the latest forensics techniques~\cite{tolosana2020deepfakes}. Current visual forgery methods can be grouped into two categories: graphics-based and feature-based. The former often mixes and disguises fake artefacts with common ones to produce a phoney image or video; for example, FaceSwap~\cite{matthewearl2020Jul} can transfer a whole face from one person to another (rather than just facial expressions, as in previous forgery methods~\cite{thies2016face2face}) in real time, even on commodity hardware. The latter method relies on the power of DNNs to increase the level of realism of visual content even further. It is often used in video forgery, such as ins DeepFake \cite{iperov2020Aug}, since the generation of fake videos requires more fine-grained and precise features than in image forgery in order to guarantee consistency between video frames~\cite{mirsky2021creation}.

In this continual war between visual forgery (measures) and visual forensics (countermeasures)~\cite{tolosana2020deepfakes,liu2017approach,gomez20183d}, the performance of each of these approaches has not been carefully compared using the same benchmarks. This is primarily due to the challenges in aligning the different settings. Countermeasures are often proposed for a previous forgery technique, and soon become obsolete for a new forgery measure. Consequently, the interpretation of performance results from forensic techniques is challenging, since baselines tend to change quickly over time. Moreover, visual forgery and visual forensics have not been subject to a fair comparison, as the reported performance is generally based on small datasets and a limited range of adverse conditions. The recent social and political damage from fake visual content requires a common ground to allow us to understand the timeline, compare variations, and keep pace with this war.

In this paper, we report the first independent dual benchmarking study to evaluate visual forgery and forensics methods in a unified framework. Using this framework, a comprehensive performance comparison is conducted on a wide range of state-of-the-art forensic techniques, forgery baselines, and real-world datasets. To envision the future of the forgery/forensics war, we also conduct in-depth analyses of synthetic datasets to extract insights on the performance behaviour of the benchmarked methods. Based on these results, we propose several guidelines for the selection of appropriate visual forensic techniques for particular application settings. Researchers and gate-keepers can use our generic framework to reduce the complexity of future benchmarking studies. 

\begin{table*}[!h]
\centering
\caption{Comparison Between Existing Benchmarks on Facial Forensics}
\label{tbl:survey_comparison}
\vspace{-1em}
\resizebox{0.9\textwidth}{!}{
\begin{tabular}{c|c|c|c|c|c|c|c|c}
\toprule
    \multirow{2}{*}{Benchmarks} &\multirow{2}{*}{\#Forensics} & \multirow{2}{*}{\#Forgery} & \multirow{2}{*}{\#Adversaries} & \multirow{2}{*}{Image} &\multirow{2}{*}{Video} & \multirow{2}{*}{Code} & \multicolumn{2}{|c}{New Datasets} \\
    \cline{8-9}
    & & & & & &  & Real & Synthetic \\
\midrule

    Our & 8 & 7 &  6 & \faCheck & \faCheck & \faCheck  & 100K\faPhoto + 19K\faVideoCamera & 1M\faPhoto + 21K\faVideoCamera\\

    Rossler~\cite{rossler2019faceforensics++} & 6 & 4 &  1 & \faCheck & \faCheck & \faCheck & 1K \faVideoCamera & 4K \faVideoCamera \\ 
  Brockschmidt~\cite{brockschmidt2019generality} & 2 & 6 & 0 & \faCheck & \faRemove & \faRemove & None & None  \\
   
   Feng~\cite{feng2020deep} & 8 & 4 & 0 & \faCheck & \faRemove & \faRemove & None &  None  \\ 
   
    Charitidis~\cite{charitidis2020face} & 3 & 3  &  1 & \faCheck & \faCheck & \faRemove & None & None  \\ 
   
    Dang~\cite{dang2020detection} & 7 & 6 & 0 & \faCheck & \faCheck & \faCheck & 59K \faPhoto\ + 1K \faVideoCamera & 240K \faPhoto\ + 3K \faVideoCamera  \\ 
   Hussain~\cite{hussain2021adversarial} & 3 & 1  & 1 & \faCheck & \faCheck & \faRemove & None & None  \\ 
   

\bottomrule
\end{tabular}
}
\vspace{-1em}
\end{table*}

The contributions of our dual benchmark can be summarised as follows:

\begin{itemize}
	
	\item \textbf{A systematic and comparative review}. We review the landscape of visual forensics and visual forgery, including the most up-to-date deep learning techniques in influential international conferences and journals, including NIPS, CVPR, Pattern Recognition, ICCV, ICLR and others. Unlike traditional surveys, our study includes new empirical comparisons between the state-of-the-art baselines. \autoref{tbl:survey_comparison} summarises the differences between our benchmark and other benchmarks in this landscape.   
	
	\item \textbf{Extensible taxonomy}. We categorise the
	most representative visual forgery approaches, including graphics-based techniques (FaceSwap-2D~\cite{matthewearl2020Jul}, FaceSwap-3D~\cite{MarekKowalski2020Jul}, 3D-Morphable face model \cite{tran2018nonlinear}) and feature-based techniques (DeepFake~\cite{iperov2020Aug}, StarGAN~\cite{choi2018stargan}, ReenactGAN~\cite{wu2018reenactgan}, Monkey-Net~\cite{siarohin2019animating} and X2Face~\cite{wiles2018x2face}). 
	Visual forensics approaches are also comprehensively categorised: (i) computer vision techniques, including frequency domain-based detectors (FDBDs) \cite{durall2019unmasking}, Visual-Artifacts \cite{matern2019exploiting}, and head pose-based detectors (HPBDs) \cite{yang2019exposing}; and (ii) neural network techniques, such as Mesonet \cite{afchar2018mesonet}, Capsule \cite{nguyen2019capsule}, XceptionNet \cite{chollet2017xception}, and GAN-fingerprint \cite{yu2019attributing}. 
	
	\item \textbf{New forgery datasets.} We apply these forgery techniques to generate forged content, resulting in a sizeable collection of datasets that is then used to explore the ability of forensics techniques to handle malicious applications. Comparing to existing datasets, our dataset covers larger range of contents and forgery types (8 techniques in 3 types, results in 1,000,000 forged images and 21,095 videos), which enables a thorough investigation of the field.  
	
	\item \textbf{Reproducible dual benchmarking}. We publish the first large-scale reproducible benchmarking framework that can assist in a dual comparison of a wide range of forensic and forgery techniques. It can also be used to justify new methods put forward in future studies~\cite{yang2018benchmark}. The framework is designed in a component-based architecture, which allows the direct integration of new forgery and forensics techniques besides the default ones. Our framework also provides the application layer, which aids the investigation of different imagery factors on the forensics performance~\footnote{\url{https://github.com/tamlhp/dfd_benchmark}}.
	
	\item \textbf{Affluent resources}. We conduct evaluations in a fair manner. We try the best to reproduce the representative techniques in the same setting while maintaining their performance reported in the original work. Our findings are reliable and reproducible, and the source code is publicly available~\footnotemark[1].  In addition to being used for comparison purposes, our framework offers reusable components that can reduce the development time.

	\item \textbf{Surprising findings}. While some of our experimental findings confirm common visual forgery studies, many of them are quite surprising, e.g. illumination factors such as brightness and contrast may significantly influences the performance while it is rarely even mentioned in visual forgery publications. Also, we have found that the current SOTA visual forensic techniques are lack of generalization ability, resulting in some difficulties with unseen forgery methods.
	
	\item \textbf{Wide range of adversarial conditions}. We simulate various different properties of the imagery content, including the brightness, contrast, noise, resolution, missing pixels and image compression. The framework is also used to visualise the effects of varying these configurable parameters. Based on both our quantitative and qualitative findings, a deeper understanding of the detection sensitivity to adverse conditions is extracted.
	
	\item \textbf{Performance guideline}. We present an exhaustive list of performance results at different levels of granularity. From this, we extract a comparative reference that can be used to select an appropriate forensic approach in particular cases of forgery.

	\item \textbf{Competency guideline}. We provide a robustness comparison of the visual forensic techniques under different adversarial conditions. This serves as a guidance for choosing forensic technique when no prior information on the forgery contents is provided. 
	
\end{itemize}

In the remainder, the paper is organised as follows. We discuss different visual forgery techniques in \autoref{sec:generators}, which is divided into two subsections that discuss graphics-based  and feature-based techniques. \autoref{sec:detectors} introduces the representative visual forensics techniques used in this benchmark. \autoref{sec:setup} introduces the setup used for our benchmark, including the component-based design, datasets, metrics and evaluation procedures. \autoref{sec:exp} reports the experimental results. \autoref{sec:con} provides a summary of the findings and practical guidelines as well as concludes the paper.

\section{Visual forgery techniques}
\label{sec:generators}

Visual forgery techniques aim to create a false image/video by injecting incorrect information (e.g. a false identity) into an original image/video. We can classify these into two categories: (i) \emph{graphics-based} approaches, such as FaceSwap-2D \cite{matthewearl2020Jul}, FaceSwap-3D~\cite{MarekKowalski2020Jul}, and 3D-morphable face models \cite{tran2018nonlinear}; and (ii) \emph{feature-based} approaches, such as DeepFake \cite{korshunov2018deepfakes}, StarGAN \cite{choi2018stargan}, ReenactGAN \cite{wu2018reenactgan} and MonkeyNet \cite{siarohin2019animating}. As shown in \autoref{tbl:forgery}, eight representative forgery techniques with their characteristics summary are used in our benchmark.

\begin{table}[!h]
\centering
\footnotesize
\vspace{-1em}
\caption{Taxonomy of visual forgery techniques}
\label{tbl:forgery}
\vspace{-1em}
\resizebox{0.9\columnwidth}{!}{
\begin{tabular}{c|c|c|c|c}
\toprule
    \multirow{2}{*}{Name} &  \multicolumn{3}{|c|}{Forgery type}  &  \multirow{2}{*}{Video specific} \\
        \cline{2-4}
    & Id Swap & Att Swap & Att mani &\\
\midrule
    Deepfake \cite{iperov2020Aug} & \faCheck &  &  &   \\
    3DMM \cite{tran2018nonlinear} &  & \faCheck &  &   \\
    FaceSwap-2D \cite{matthewearl2020Jul} & \faCheck &  &  &   \\
    FaceSwap-3D \cite{MarekKowalski2020Jul}& \faCheck &  &  &   \\
    MonkeyNet \cite{siarohin2019animating} &  & \faCheck &  & \faCheck  \\
    ReenactGAN \cite{wu2018reenactgan} & \faCheck &  \faCheck &  &   \\
    StarGAN \cite{choi2018stargan} &  & \faCheck & \faCheck &   \\
    X2Face \cite{wiles2018x2face} & &  &  \faCheck &  \faCheck \\
\bottomrule
\end{tabular}
}
\vspace{-1em}
\end{table}

\subsection{Graphics-based techniques}
\label{sec:graphic-based generator}

These techniques are often used to replace the face of a source person (A) with the face of a target person (B) using handcrafted features (e.g. the landmark points of a human face) to forge the image. We describe several typical graphics-based techniques below.

\sstitle{FaceSwap-2D} This technique is a variant of FaceSwap~\cite{matthewearl2020Jul}, in which the \textit{colour adjustment} step is performed using histogram matching of the two images, $x_A$ and $x_B$, and a set of 68 landmark points on a 2D scale is then used to fit the face of the target person B onto the source image of person A. 

\textit{Colour adjustment:} In this routine, the aim is to transfer a histogram of the image of person B in order to match it with the histogram for the image of person A. This process involves the following steps:
1) Calculate the histograms $h_A, h_B$ for the input images $x_A, x_B$.
2) Calculate the normal histogram function for each image $p_A, p_B$ from $p_A(i) = \frac{h_A(i)}{N_A},p_B(i) = \frac{h_B(i)}{N_B}$, where $N_A,N_B$ are the number of pixels in each image.
3) Calculate the cumulative distribution function of each image:
$c_A(i) = \sum_{j=0}^{i}p_A(j), c_B(i) = \sum_{j=0}^{i}p_B(j)$.
4) For each grey level $k$ of image A, find the grey level $j$ from the formula:
$j = argmin_{v} |c_B(k) - c_A(v)|$.

\textit{Head poses matching:} After colour calibration, the face of the target person in $x_B$ is extracted and fitted onto $x_A$ using a 
linear transformation matrix $\Omega$ guided by the 68-point landmark system in the 2D setting: 
\begin{equation}
\label{eqn:fs_2D}
\footnotesize
\Omega^* = arg \min_\Omega \sum_{i=1}^{68} ||\Omega p_i^T - q_i^T||^2
\end{equation}
where $p_i^T, q_i^T$ are the vector landmark coordinates of the landmark points in $x_A$, $x_B$ respectively, and $\Omega^*$ is the optimised value of the transformation matrix as shown in \autoref{eqn:fs_2D}. This optimisation is often referred to as the orthogonal Procrustes problem~\cite{schonemann1966generalized}, for which the direct solution can be obtained as follows:
$
\footnotesize
\Omega^* = UV^T
$
where $USV^T = P^TQ$ with $Q,P$ are the landmark matrices of $x_A, x_B$ and $USV^T$ is a single value decomposition from $P^TQ$.

\sstitle{Faceswap-3D}
This graphics-based technique goes beyond the Faceswap-2D method by considering the landmark points in three dimensions when matching the head poses \cite{MarekKowalski2020Jul}. 
The key idea in this approach is the 3D setting of facial landmarks, which makes the generated image harder to detect:
\begin{equation}
\footnotesize
\begin{bmatrix}
        X \\
        Y\\
        Z
    \end{bmatrix} = R\begin{bmatrix}
        U \\
        V\\
        W
    \end{bmatrix}+t
\end{equation}
where $[U,V,W]^T$ are the coordinates of the facial landmark points of a standard face, $[X,Y,Z]^T$ are the camera coordinates, $R$ is a $3\times3$ rotation matrix, and $t$ is a $3\times1$ translation vector. After 3D modelling, the projection into 2D is carried out as follows:
\begin{equation}
\label{eqn:proj_3D}
\footnotesize
\begin{bmatrix}
        x\\
        y\\
        1
    \end{bmatrix} = s \begin{bmatrix}
        f_x & 0 & c_x \\
        0 & f_y & c_y \\
        0 & 0 & 1
    \end{bmatrix}*\begin{bmatrix}
        X \\
        Y \\
        Z
    \end{bmatrix}
\end{equation}
where $f_x$ and $f_y$ are the focal lengths in the X and Y dimensions, $s$ is a scaling factor, and $(c_x,c_y)$ is the optical centre.

After 3D modelling, 
the projection of the 3D landmarks onto 2D equivalents, as shown in \autoref{eqn:proj_3D} should match the 2D landmarks in the image. This can be formulated as an optimisation problem:
\begin{equation}
\footnotesize
\min_{R,t,s} = \sum_{i=1}^n \Bigg|\Bigg|  \begin{bmatrix}
        x_i \\
        y_i \\
        1
    \end{bmatrix} - s \begin{bmatrix}
        f_x & 0 & c_x \\
        0 & f_y & c_y \\
        0 & 0 & 1
    \end{bmatrix} \Bigg( R \begin{bmatrix}
        U_i \\
        V_i \\
        W_i
    \end{bmatrix}+t \Bigg) \Bigg|\Bigg|^2
\end{equation}
This optimisation is often referred as a P3P problem, and a strategy for solving this can be found in \cite{quan1999linear}. The output is a tuple $(s, R, t)$, called the estimated head pose. To swap the face of the target person in $x_B$ with the face from the source image $x_A$, the face texture of B is injected with the estimated head pose $(s_A, R_A, t_A)$ of $x_A$.

\sstitle{3D-Morphable face model (3DMM)}
This is a graphics-based technique that is also used to model a person's face in 3D with shapes and textures~\cite{tran2018nonlinear}. However, instead of using a linear mapping  from 3D to 2D, this model uses a nonlinear mapping that is learned by an encoder-decoder deep neural network. 

Formally, given a set of 2D face images $ {I_i }_{i=1}^N$, 3DMM constructs three deep neural networks: (i) an \textit{encoder} $ E: I\rightarrow \{m,f_S,f_T\} $ that learns the
projection parameter $m$, the shape parameter $f_S \in \mathbb{R}^s$ and the texture parameter $f_T \in \mathbb{R}^t$; (ii) a \textit{shape decoder} $D_S: f_S \rightarrow S$; and (iii) a \textit{texture decoder} $D_T: f_T \rightarrow T$ that reconstructs the 2D shape $f_S$ and features $f_T$ to create a 3D shape $S$ and a texture $T$ respectively. The three components $E, D_S, D_T$ are trained simultaneously, to minimise the reconstruction loss of the input face image:
$
\footnotesize
L_{rec} = \sum_{i=1}^N \big\| R( E_M(I_i), D_S(E_S(I_i)), D_T(E_T(I_i)) )  - I_i \big\|_1
$
The reconstruction loss can also be combined with a landmark loss $L_L$ (a geometric constraint) and an adversarial loss $L_{adv}$ (to ensure realistic rendering):
$
\footnotesize
L = L_{rec} + \lambda_{adv}L_{adv} + \lambda_LL_L
$
to enhance the quality of the training process. 
After the training process, 
the fake image can be generated by injecting the face texture of the target person $f_T^B$ with the projection and shape parameters of the source image $\{ m^A, f_S^A \}$.

\subsection{Feature-based techniques}

Recent fake image generators have leveraged advanced neural network architectures such as generative adversarial networks (GANs) and variational autoencoders (VAEs)~\cite{kingma2013auto} to produce forged images of superior quality, without the need for feature engineering or expert knowledge. 
We describe some representative feature-based techniques below.

\sstitle{DeepFake}
This is an autoencoder-based model~\cite{iperov2020Aug} that can replace one person's face with any other faces. The typical architecture of this model is composed of one encoder $En$ and two decoders $De_{A}$, $De_{B}$ (for the source and target person). 
The two encoder-decoder pairs ($En, De_{A}$) and ($En, De_{B}$) are trained separately, using the \textit{reconstruction loss}: 
$
\footnotesize
L_{rec}^{X} = En_{x_{X}} [ ||x_X - De_{X}(En(x_X))||_1 ]
$
where $||.||_1$ denotes the L1 norm and $X \in \{A, B\}$. The aim of the loss function is to guarantee that the decoders $De_A$ and $De_B$ can accurately reconstruct the original images from the encoded features. 

To enhance the counterfeit, recent DeepFake variants (e.g. Faceswap-GAN~\cite{shaoanlu}) were inspired by the GAN model to 
add two discriminators $D_A$ and $D_B$, which separate fake from real images with an additional \textit{adversarial loss}: 
$
\footnotesize
    \label{eqn:adv_loss}
    L_{adv} = E_{x_X}[\log D(x_X)] + E_{x_X}[\log(1 - D_X(De_X(En(x_X)))]   
$
The loss functions used to train the ED and D networks are:
\begin{align}
\footnotesize
L_{ED} &=  L_{rec} + \lambda_{adv} L_{adv} \\
L_{D} &=  -L_{adv} 
\end{align}
where $\lambda_{adv}$ is a balancing hyper-parameter between originality (reconstruction loss) and realistic rendering (adversarial loss). After the training process, the fake image can be obtained by applying the decoder $De_B$ to the encoded feature of the input image of A: $x_f^{B \rightarrow A} = De_{B}(En(x_A))$.



\sstitle{StarGAN}
This is a GAN-based model~\cite{choi2018stargan} that can generate a fake image by manipulating the facial attributes (e.g. hair color, skin, gender, facial expression). To achieve this, StarGAN first groups the training images that share a particular combination of attributes as a \textit{domain}. It then uses a generator \textit{G} to learn a mapping between multiple domains: $G(x,c) \rightarrow y$, where $x$ and $y$ are the input and output images, respectively, and $c$ is a target domain which is randomised in the training process to enable a flexible transition. The model also employs a discriminator $D$ to classify the image as real or fake ($D_{src}(x)$), and to identify the domain to which the image belongs ($D_{cls}(x)$).
To efficiently train $G$ and $D$ in an adversarial way, StarGAN uses the following losses: 

\begin{itemize}
	\item \textit{Adversarial loss:} This loss function aims to ensure that the generated image is indistinguishable from the real image:  
$
\footnotesize
    L_{adv} = E_{x}[log(D_{src}(x))] + E_{x,c}[log(1 - D_{src}(G(x,c)))]   
$
In other words, the generator $G$ minimises the entropy of $D_{src}$, while the discriminator $D$ maximises it.

\item \textit{Domain classification loss:} StarGAN uses two separate domain classification losses:
\begin{align}
\footnotesize
    L_{cls}^r &= E_{x,c'}[-log(D_{cls}(c' | x))] \\
L_{cls}^f &= E_{x,c'}[-log(D_{cls}(c' | G(x,c)))]
\end{align}
where $L^r_{cls}$ is used for real images and $L^f_{cls}$ is for fake images. This helps both $D$ and $G$ to capture well the domain information of the images more effectively. 

\item \textit{Reconstruction loss:} This guarantees that $G$ translates only the domain information from the original image:
$
\footnotesize
L_{rec} = E_{ x,c,c'}[||x-G(G(x,c),c')||_1]
$
The cycle consistency principle~\cite{kim2017learning} is applied to guarantee that the generator can reconstruct the original image using the domain information.
\end{itemize}


\sstitle{ReenactGAN}
This is another GAN-based model~\cite{wu2018reenactgan} that can transfer the facial movements and expressions to generate a fake image. Instead of using a pixel-wise transformation, the model maps the target image onto a latent space that closely captures the facial contours (i.e. boundaries). 

The architecture of ReenactGAN consists of three DNNs: (i) an \textit{encoder} ($En$), which embeds the target image into a latent boundary space; (ii) a \textit{target-specific decoder} ($De$), which converts the embedding in latent space to the source image $x_A$; and (iii) a \textit{boundary transformer} $\phi$, which fits the boundaries of the target face to those of the source image. The encoder ($En$) and decoder ($De$) are designed using deep networks (e.g. VGG-16) inspired by Pix2Pix \cite{isola2017image}, and are trained using the combined loss function: 
$
\footnotesize
L(En,De) = L_{adv} + L_{rec} + L_{feat}
$
where $L_{adv}$ is an adversarial loss similar to that used in \textit{DeepFake} (see \autoref{eqn:adv_loss}), which acts as a discriminator between the real sample $x$ and the reconstructed sample $De(En(x))$. $L_{rec}$ is the L1 reconstruction loss, which guarantees that the encoder $En$ only encodes the boundary, and $L_{feat}$ normalises the hidden features of the network. The decoder $De$ is trained on the target image $x_B$ rather than the source image $x_A$. The \textit{boundary transformer} $\phi$ is trained to reconcile the boundary spaces of the two images:
$
\footnotesize
L(\phi,De) = L_{cycle} + L_{adv} + L_{shape}
$
where $L_{cycle}$ is a cycle consistency loss for the boundary transformer $\phi$, $L_{adv}$ is an adversarial loss for $\phi$, and $ L_{shape}$ constraints the shapes of the transformed boundaries to be similar to those of the source.

\sstitle{Monkey-Net}
This is a motion-driven GAN-based model~\cite{siarohin2019animating} that can inject an action by a target person B into a source image $x_A$ without changing the other properties. The model first learns a set of motion-specific keypoints in an unsupervised manner, which allow it to describe relative movements between pixels. Then, only the relevant motion-specific patterns of the source image $x_A$ are transferred to $x_B$ while the other information is unchanged. 

The Monkey-Net framework contains three components. The first is the \textit{keypoint petector}, denoted by $\Delta$, which extracts motion-specific keypoints from the input images $x_A$ and $x_B$. The output of the module is fed to the second component, the \textit{dense motion predictor}, which translates the sparse key points into a motion heat map. The third module is called the \textit{motion transfer} network, and this combines the dense motion heat map with the source image $x_A$ to produce the fake image. To train the model, a generator network $G$ is trained together with the \textit{keypoint detector} $\Delta$, such that $G$ can reconstruct $x_A$ given $\Delta(x_A)$, $\Delta(x_B)$ and $x_B$, while the discriminator $D$ is responsible for distinguishing the real image from the fake one, as follows:
\begin{align}
\footnotesize
    L_{adv}^{G} &= E[(D(x' \oplus \Delta(x_B))-1)^2)] \\
    L_{adv}^{D} &= E[(D(x_A \oplus \Delta(x_B))-1)^2] + E[D^2(x' \oplus \Delta(x_B))]
\end{align} 
where $x'$ is the generated fake image and $\oplus$ denotes the concatenation along the channel axis.

\sstitle{X2Face} This is a lightweight self-supervised neural network model \cite{wiles2018x2face} that can manipulates the pose and expression of a given face image. X2Face takes two inputs: a source frame and a driving frame. The source frame is forward to a convolutional network (inspired from U-net \cite{wiles2018x2face}) called \textit{embedding network}, which learns a bilinear sampler to construct the mapping from the source frame to an embedded face. The driving frame is put through an encoder-decoder architecture named \textit{driving network}, which learns a bilinear sampler to transform the embedded face to the generated frame. 

The network is trained in two stages. The first training stage is fully self-supervised, which uses the images sampled from the same video. To this end, the generated frame and the driving frame have the same identity, which guarantee latent embedding learnt from \textit{driving network} must encode variation factors (e.g. pose, expression, zoom) by a pixelwise L1 loss between the generated and the driving frames. In the second training stage, additional identity loss functions are applied to enforce that the identity of the generated and the source frames are the same. To this end, the trained network is able to inject into a given source frame variation factors from a driving frame of a different person while maintain his identity. Further details can be found in \cite{wiles2018x2face}.

\section{Visual forensics techniques}
\label{sec:detectors}

Following the rapid development of forgery techniques as well as the emerging threat of forged artefacts, many studies of visual forensics methods have been carried out. We can divide these methods into two categories: (i) \textit{computer vision} techniques, which rely on handcrafted features to detect the anomalous patterns (e.g. frequency, head pose); and (ii) \textit{deep learning} techniques, which leverage the advances in deep learning to automatically learn hidden features that are non-trivial for humans. 

\subsection{Computer vision techniques}

\sstitle{FDBD}
In this approach~\cite{durall2019unmasking}, the frequency characteristics of the input image are investigated to discover anomalous content. A frequency domain analysis is used to exploit 
the repetitive nature of the frequency characteristics of images~\cite{pan2012exposing}. 

More precisely, FDBD adopts a discrete Fourier transform (DFT) to decompose the input image into sinusoidal components of various frequencies. This spectral decomposition of the input image (which is treated as an $M \times N$ signal) reveals the distribution of signal energy over different frequency ranges:
$
\footnotesize
X_{k,l}=\sum_{n=0}^{N-1} \sum_{m=0}^{M-1} x_{n,m} . e^{-\frac{i2\pi}{N}kn} . e^{-\frac{i2\pi}{M}lm}
$
where $X_{k,l}$ is the frequency-domain representation, in which each frequency is associated with a signal amplitude and a phase.

\sstitle{Visual-Artifacts}
This is a fake image detection technique~\cite{matern2019exploiting} that relies on several visual features~\cite{krestenitis2020recurrent,wei2020minimum} (called artefacts) that emerge as part of the processing pipeline of common facial manipulation techniques (e.g. DeepFake and FaceSwap). 

\textit{Global Consistency}:
Fake image generators, and especially \textit{feature-based} techniques, often smooths a given face by interpolating the latent space of network features with supporting data points. However, these data points are not necessarily meaningful when new faces are generated, resulting in a mixture of different facial characteristics (e.g. differences in colour between the left and right eyes), which is referred to as global consistency.

\textit{Illumination estimation}:
An original image may contain incident illumination, and this poses a challenge when rendering a fake image with similar illumination conditions. Visual forgery techniques often leave traces of illumination-related artefacts: for example, a typical artefact of the DeepFake algorithm is a shading effect around the nose, in which one side is too dark. 
 
\textit{Geometry estimation}:
Facial geometry is often taken into account in graphics-based models (e.g. 3D-Morphable) or feature-based generators (e.g. geometry estimators) to make the counterfeit image more realistic. However, this is often approximate, and leads to inaccurate details (artefacts). These artefacts typically appear along the boundary of the face mask (e.g. the nose, eyebrows and teeth) in the form of \textit{blending} spots (strong edges or high contrast) or \textit{holes} (missing detail).

\sstitle{HPBD} 
This approach~\cite{yang2019exposing} exploits the head pose limitations of forgery models. When a visual forgery technique is used to inject the face of the target person into the source image, the facial landmarks may be mismatched. These errors in landmark locations can be discovered using a 2D head pose estimation between the real and fake regions of the input image. To achieve this, HPBD compares head poses across all facial landmarks and uses the central region to look for anomalies and discrepancies. 

More precisely, the model utilises the 3D configuration of the facial landmarks, as described in \textit{Faceswap-3D} (see \autoref{sec:graphic-based generator}) to estimate the head pose configuration $(s, R, t)$. Note that HPBD splits a system of 68-landmark points into two parts, representing the central and border regions. As a result, the estimation produces two different rotation matrices $R$, where $R^T_a$ estimates the head pose of the whole face, and $R^T_c$ estimates the head pose of the central region landmarks. The vectors $\vec{v}_a$ and $\vec{v}_c$ representing the orientations of the head are then calculated by $\vec{v}_a=R^T_a\vec{w}$ and $\vec{v}_c=R^T_c\vec{w}$, where $\vec{w}=[0,0,1]^T$ are the world coordinates. Finally, the level of inconsistency between $\vec{v}_a$ and $\vec{v}_c$ is measured using the cosine distance. Empirical results show that this value is small for real images, and significantly larger for synthesised images. This feature is therefore a robust indicator for use in separating fake images from real ones.

\subsection{Deep learning techniques}

\sstitle{Mesonet}
This is a deep learning fake image detection technique~\cite{afchar2018mesonet} that detects forged images at a mesoscopic level of analysis. Two variants of it have been proposed based on the mesoscopic properties of the image, namely \textit{Meso-4} and \textit{MesoInception-4}.

\textit{Meso-4:} This variant is designed with four layers, which alternate between convolution and pooling layers. Each convolutional operation is combined with ReLU activation~\cite{dahl2013improving} and batch normalisation~\cite{ioffe2015batch} for better generalisation.

\textit{MesoInception-4:} This variant improves \textit{Meso-4} through the use of two inception modules~\cite{afchar2018mesonet} instead of the first two convolutional layers. The idea behind the inception operation is to enrich the function space of the model by applying different kernel shapes to multiple convolutional layers simultaneously. The other vanilla convolution layers in \textit{Meso-4} are replaced by dilated convolutions \cite{shi2017single} to avoid overfitting. 

\sstitle{Capsule}
This is an NN model~\cite{nguyen2019capsule} that is capable of detecting various kinds of manipulated images, ranging from replay attacks (using a printed photo in front of a webcam or camera) \cite{chingovska2012effectiveness, de2013can} to sophisticated CNN-based techniques. The model leverages the ``capsule'' architecture~\cite{nguyen2019capsule}, and combines it with expectation maximisation and dynamic routing ~\cite{sabour2017dynamic, hinton2018matrix} to capture the hierarchical relationships between different poses~\cite{nguyen2019capsule}. This significantly improves the performance of counterfeit detection, especially against highly realistic photos/videos.

The model first locates the face in the image and rescales it to a size of $128 \times 128$. This is then forwarded to a VGG-19 network~\cite{simonyan2014very}. Next, the output latent features are fed to a capsule network that contains: (i) three primary capsules, each of which integrate statistical pooling to enhance forgery detection; 
and (ii) two output capsules, which are dynamically routed from the three previous capsules. The model is trained using a cross-entropy loss function:
$
\footnotesize
    L = - y log\hat{y} + (1 - y) log(1 - \hat{y})
$
where $y$ is the predicted label and $\hat{y}$ is the ground-truth.

\sstitle{XceptionNet}
This approach~\cite{chollet2017xception} adopts the Inception architecture \cite{szegedy2016rethinking, szegedy2017inception} to extract the underlying features of input images to distinguish between fake and real images. The original Inception architecture maps the input data from the original space to multiple smaller spaces separately, and the cross-channel correlations between smaller spaces are then put together via convolutional layers. 

XceptionNet goes beyond existing Inception architectures by entirely decoupling the correlations across space and channels.  It has 36 convolutional layers, which act as the feature extraction module of the whole network. This module in turn consists of three parts, each of which is constructed from a linear stack of
depth-wise separable convolution layers with residual connections. This linear stacking increases the flexibility of development in terms of implementation and modification 
for high-level libraries such as Keras. The first part, referred to as the \textit{entry flow}, processes the data once, while the second, called the \textit{middle flow}, processes the data eight times. The final part, called the \textit{exit flow}, then processes the data once. Finally, a logistic regression layer is applied for binary classification (real/fake). 

\sstitle{GAN-fingerprint}
This model~\cite{yu2019attributing} detects forged images by tracing the fingerprints of GAN~\cite{marra2018detection, mo2018fake}, which forms the heart of visual forgery algorithms such as DeepFake. Two types of fingerprint are investigated: \textit{model fingerprints} and \textit{image fingerprints}. 

\begin{itemize}
	\item \textit{Model fingerprint:} This approach is based on the observation that even if two well-trained GAN models vary in terms of the hyper-parameter configurations, the non-convexity of the loss functions and the adversarial equilibrium between the generator and discriminator, their high-equality generation is equivalent. This uniqueness can be exploited to trace GAN-based modifications. 

	\item \textit{Image fingerprint:} 
If fake images are generated by the same GAN instance, they often have stable, common patterns, and vice-versa. This uniqueness hints that the encoding of an image fingerprint is possible. 
\end{itemize}

Using these two observations, GAN-fingerprint learns the model fingerprint for each source, and then uses it to map an input image to its fingerprint. 
Formally, given an image-model pair $(I,y)$ where $I$ is the input image and $y \in \mathbb{Y}$ is a GAN instance, the model learns a reconstruction function $R: I \rightarrow R(I)$ using the following pixel-wise reconstruction losses: 
\begin{align}
\footnotesize
L_{pix}(I) &= ||R(I)-I||_1 \\
L_{adv}(I) &= D_{rec}(R(I))-D_{rec}(I) + c(R(I),I|D_{rec})
\end{align}
where $D_{rec}$ is a discriminator and $c$ is a gradient penalty regularisation term \cite{gulrajani2017improved}. The image fingerprint $F^I_{im}$ then is calculated as the reconstruction residual: $F^I_{im} = R(I) - I$. The model fingerprint $F^y_{mod}$ is defined using freely trainable parameters with the same size as $F^I_{im}$. The model then maximises the correlation $corr$ between $F^y_{mod}$ and $F^I_{im}$ over the instance set $\mathbb{Y}$, using the cross-entropy loss:
$
\footnotesize
L_{cls}(I, y) = -log \frac{cor({F^y_{mod}, F^I_{im})}}{\sum_{\hat{y} \in \mathbb{Y}}{cor(F^{\hat{y}}_{mod}, F^I_{im})}}.
$
The losses $L_{adv}$, $L_{pix}$ and $L_{cls}$ then are put together with a weighted-sum combination to train the model.

\section{Dual benchmarking framework}
\label{sec:setup}

In this section, we introduce our dual benchmarking framework, including datasets, measurements, and experimental procedures.

\subsection{Datasets}

One of our contributions is a dual benchmarking dataset, which includes forged contents generated by various visual forgery techniques introduced in section \autoref{sec:generators}. Although there have been existing forgery datasets \cite{deepfakeinthewild2020Jul, li2020celeb, dolhansky2020deepfake}, our dataset contains greater diversity by using larger pool of forgery techniques and cover more manipulation types. This allows us to analyse the performance of the forensic techniques on various types of forgery.

\subsubsection{Dual-benchmarking datasets (DBD)} 
\begin{figure}[!h]
    \centering
    \vspace{-.5em}
    \includegraphics[width=0.55\linewidth]{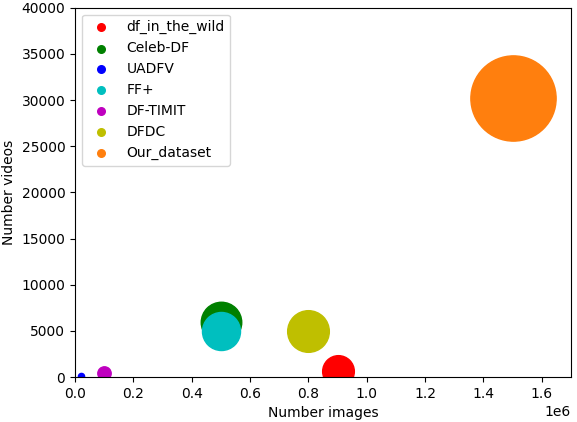}
    \vspace{-1em}
    \caption{Size of facial forgery datasets}
    \label{fig:dataset_size}
    \vspace{-1em}
\end{figure}

We sample the real face images and videos from the facial datasets CelebA-HQ~\cite{karras2017progressive}, DFDC~\cite{karras2017progressive} and FaceForensic++~\cite{rossler2019faceforensics++} since the images cover large variation of gender, age, expression and quality. Our final dataset contains 100,000 source images and 19,000 source videos. 48.3\% of images and video frames are from male subjects and 51.7\% of them are from females; and the majority of samples cover the range of age from 18 to 50. 
The \autoref{fig:dataset_size} illustrates the size of our dataset comparing to other popular facial datasets such as FaceForensic++, DF-TIMIT, UADFV and DFDC.   

\sstitle{Identity swap} For identity swap, we adopt the methods FaceSwap-2D, FaceSwap-3D, 3DMM and DeepFake. For FaceSwap-2D and FaceSwap-3D, we randomly choose pair of images from different person in our real image pool as inputs (source and target images). We apply the default landmark system for both techniques, results in 82,590 fake images for FaceSwap-2D and 50,378 fake images for FaceSwap-3D. With 3DMM, we use the pretrained model provided by the authors to analyse the input images and synthesise 91,885 forged images. For DeepFake, we choose and train the model for 5 celebrities Emmanuel Macron, Kathleen, Jack Ma, Theresa May, Donald Trump to inject a face from one image into the other image. We also reuse the processed videos from FaceForensic++, which contains 3,000 forged videos generated from 1,000 real YouTube videos. 

\sstitle{Expression swap} For the swap of expression from one facial image/video to the other, we apply the X2Face, Monkey-Net and ReenactGAN methods. For Monkey-Net, we leverage the default video driver provided by the authors to produce the forged contents. For X2Face, we use the images sampled from the same video for the self-supervised training stage and choose the frames from another video as driving frame for the second training stage. With ReenactGAN, we apply the full architecture including a pretrained encoder, a self-trained transformer and a decoder to guarantee the best performance. In total, we generated 11,095 videos and 66,509 images for this category. 

\sstitle{Attribute manipulation} We leverage the StarGAN technique, a GAN-based image-to-image translation method, for the swap of facial attributes such as gender, skin color, hair and beard. For each sampled real image, we generate 8 fake images using pretrained attribute combination, results in 299,224 fake images. 

\subsubsection{External datasets} We employ the following external dataset in our assessment of the visual forensics techniques: 

\begin{itemize}
\item \emph{DeepFake-in-the-wild:} This consists of 7,314 face sequences extracted from 707 DeepFake videos that were collected entirely from the internet~\cite{deepfakeinthewild2020Jul}. 

\item \emph{Celeb-DF:} This contains 590 real videos of short interviews of 59 celebrities of different genders, ages and ethnic groups. From these real videos, 5,639 fake videos were forged using DeepFake~\cite{li2020celeb}.

\item \emph{DFDC:} This contains 5,244 videos. The actors were crowdsourced and are diverse in terms of gender, skin tone and age. The background, lighting conditions and head poses are also flexible~\cite{dolhansky2020deepfake}. 

\item \emph{UADFV:} This contains 49 real videos, which are used to create 49 DeepFake videos. The average length of these videos is 11.14s, with a typical resolution of $294$ to $500$ pixels~\cite{yang2019exposing}.

\item \emph{DF-TIMIT:} This contains 10 original videos for each of 43 subjects, captured in a controlled environment and with the actors facing the camera. Forged videos are generated using a GAN-based face-swapping algorithm~\cite{mirsky2021creation}.

\item \emph{FaceForensics++ (FF++):} This includes 1000 pairs of real and synthetic videos, in which the latter are generated using Face2Face (FF1), FaceSwap (FF2), DeepFakes (FF3), and NeuralTextures (FF4)~\cite{rossler2019faceforensics++}.
\end{itemize}

\autoref{tbl:dataset_summary} summarizes the datasets using in our benchmark. Comparing to the existing datasets, our constructed dataset DBD contains larger amount of real and forged images and videos, as we utilize more forgery techniques. Also, we cover various forgery types, including the attribute manipulation, while most of existing datasets focus on identity swap. 

\begin{table}[!h]
\centering
\vspace{-1em}
\caption{Statistic of real datasets}
\label{tbl:dataset_summary}
\vspace{-1em}
\resizebox{0.9\columnwidth}{!}{
\begin{tabular}{|c|c|c|c|c|c|c|c|}
    \hline 
    \multirow{2}{*}{Dataset} & \multicolumn{2}{|c|}{Image} & \multicolumn{2}{|c|}{Video} & \multicolumn{3}{|c|}{Forgery type} \\
    \cline{2-8}
    & Real & Fake & Real & Fake & Id Swap & Att Swap & Att mani \\
    \hline
   df\_in\_the\_wild \cite{deepfakeinthewild2020Jul} & 331867 & 582561 & 70 & 707 & \checkmark & &  \\
    \hline
    Celeb-DF \cite{li2020celeb} & 71817 & 495087 & 590 & 5639 & & \checkmark &  \\
    \hline
    DFDC \cite{dolhansky2020deepfake} & 102231 & 671285 & 1131 & 4113 & \checkmark & & \\
    \hline
    UADFV \cite{yang2019exposing} & 9374 & 9358 & 49 & 49 & \checkmark & &  \\
    \hline
     DF-TIMIT \cite{korshunov2018deepfakes} & 66584 & 28994 & 430 & 640  & \checkmark & & \\
    \hline
   FF++ \cite{rossler2019faceforensics++} &  81512 & 28481 & 1000 & 4000 & \checkmark & \checkmark &  \\
    \hline
    DBD(Our\_dataset) & 100000 & 1000000 & 19000 & 21095 & \checkmark & \checkmark & \checkmark \\
    \hline
\end{tabular}
}
\vspace{-1em}
\end{table}

\subsection{Measurements}

To ensure effectiveness, we use the four metrics of precision, recall, F1-score, and accuracy, since forgery detection can be treated as a binary classification.
In terms of efficiency, a forensic method also needs to handle a large amount of data, and this is especially true for videos. The detection speed is therefore an important aspect. We measure the detection speed as the number of frames or images a technique can process per second.

\subsection{Experimental Procedures}
\label{sec:eval_proc}

We carried out in-depth performance analyses of these forensic techniques in various scenarios.

\sstitle{Dual benchmarking} 
Existing benchmarks mostly focus on a single side of the problem by comparing only visual forensics techniques against each other. In our benchmark, we also evaluate visual forgery techniques against visual forensics ones via our constructed dual benchmarking dataset. The dual analysis can provide users with an understanding of the robustness of each visual forensics technique against state-of-the-art visual forgery mechanisms. 

\sstitle{Forensic generalization and forgery feature overlapping} One of the important property of the forensic techniques is the ability of generalization. An ideal forensic technique should be capable of capturing key features and handle multiple forgery techniques, even the unseen ones. To this end, we investigate the generalization capability of the forensic techniques by training their model with forged images from one technique and testing on images from another technique. The performance of the forensic models against the new forgery technique also indicates how well the features extracted for the original technique overlap with the unseen technique.


\sstitle{Qualitative study of forgery-forensic duel} To learn how the forensic techniques predict the manipulated region or pixels from the forged contents, we apply the attention-based layer \cite{dang2020detection} to the neural network based forensic techniques such as XceptionNet. After training, the intensity of each pixel in the attention map depicts the probability of the input image's pixel being a fake region. This attention map thus helps us to reveal the suspicious spots for each type of forgery.  

\sstitle{Influence of contrast} 
Contrast is an important property of an image that can affect the performance of the visual forensic techniques. 
To investigate the impact of this factor, we changed the contrast of the images using the formula: 
\begin{equation}
\footnotesize
x = \begin{cases}
  avg + \psi (x-avg), & \text{if } 0<avg + \psi(x-avg)<255
\\
  255, & \text{if } avg + \psi (x-avg)>255
\\
  0, & \text{if } avg + \psi (x-avg)<0
\end{cases}
\end{equation}
where $ x $ is value of pixel in image, $avg$ is average of value of all pixel in image, and $\psi$ is contrast factor. 
We evaluated the performance of the visual forensic models using different contrast factors of $\{0.5, 0.75, 1, 1.5, 2\}$.

\sstitle{Effects of brightness} 
Brightness is another important property of images that can affect the performance of the visual forensic techniques, 
e.g. due to overexposure effects. We simulated a diverse range of values for brightness using a brightness factor. More precisely, we multiplied a pixel intensity by a brightness factor $\tau$ and then clipped the scaled value to a valid range of $[0, 255]$: 
\begin{equation}
\footnotesize
x = \begin{cases}
  \tau  x, & \text{if } \tau  x <255
\\
  255, & \text{if } \tau  x >255
\end{cases}
\end{equation}
After some preliminary studies, we selected five brightness factors $\{0.5, 0.75, 1, 1.5, 2\}$ that could create significant differences between images.

\sstitle{Robustness against noises} 
In our benchmark, we also studied the effects of noise on forensic performance. We simulated noise by adding a Gaussian noise $\mathcal{N}(0,\sigma)$ to the original and forged images. The intensity values were again normalised to the range $[0, 1]$, by dividing the noise-modified values by 255. The $\sigma$ value varied between $0$ and $0.25$, with a step size of $0.05$. 

\sstitle{Robustness against image resolution} 
Image resolution is another important property of images that we aimed to explore using our benchmark. It is intuitively obvious that visual content with high resolution is easier to analyse carefully than lower-resolution content. We simulated this property by resizing the original dimensions of the image ($256 \times 256$) in our dual-benchmarking dataset to $128 \times 128$, $64 \times 64$, $32 \times 32$ and $16 \times 16$, respectively. 

\sstitle{Influence of missing information} 
Another important factor is missing data. We simulated this effect by removing pixel areas from real and fake images. In fact, this was equivalent to partial occlusion, as the pixel values of this area were set to zero. We varied the sizes of the missing areas from $0\%$ to $50\%$ of the image, with a step size of $10\%$.

\sstitle{Adaptivity to image compression} 
In practice, images may be saved in compression formats, which affects the forensic performance. To simulate this compression factor, we encoded the images with the JPEG algorithm with six different values of quality: $\{50, 60, 70, 80, 90, 100\}$. 


\subsection{Reproducibility Environment}

All experiments were conducted on an AMD Ryzen ThreadRipper 3.8 GHz system with 128 GB RAM and four RTX 2080Ti graphics cards. To mitigate the randomness, we averaged the results over 50 runs for each facial forgery dataset. All algorithms were evaluated using the same standards (implemented in Python and tested on the same configuration) in order to guarantee fairness.
\section{Evaluation Reports}
\label{sec:exp}

We applied our benchmark to the aforementioned visual forensic techniques and forgery methods. Empirical results are reported below.

\subsection{Efficiency comparison}
 
 The experiment results for detection speed are shown in \autoref{tbl:model_size}. Capsule and XceptionNet were the two fastest techniques, and could process 10,000 images within 26 and 23 seconds, respectively. The processing speeds of traditional techniques such as HPBD and Visual-Artifacts were much slower, with detection times of 2,400 and 9,100s per 10,000 images. This is because these techniques require a large amount of time to extract the handcrafted features from the input images. Note that the model sizes for the traditional techniques are not reported, since they do not use deep neural networks. 
 
\begin{table}[H]
\centering
\vspace{-1em}
\caption{Model size and detection speed}
\label{tbl:model_size}
\footnotesize
\vspace{-1em}
\resizebox{0.6\columnwidth}{!}{
\begin{tabular}{c|c|c}
\toprule
     & \#Parameters &
     Detection Speed (s/10,000 images)  \\
\midrule
    Mesonet4& 28073 & 57\\
   Capsule & 3895998 & 26\\
    XceptionNet &  21861673 & 23 \\
    GAN-fp & 14252563 & 243\\
    FDBD & N/A & 245 \\
    HPBD & N/A & 2400\\
    VA & N/A & 9100\\
\bottomrule
\end{tabular}
}
\vspace{-1em}
\end{table}

\subsection{End-to-end comparison}

In this experiment, we examined the end-to-end performance of visual forensics techniques on existing datasets. 
The results are shown in \autoref{tbl:accuracy_datasets}. Xception performed the best, with the highest accuracy on all datasets except UADFV. Xception also achieved high values of precision, recall and F1-score for all the datasets, with results ranging from 0.75 to 1. For the UADFV dataset, GAN-fingerprint outperformed Xception on all three metrics, and even reached absolute results. This is because UADFV is a small dataset using only the DeepFake technique. For the other datasets, GAN-fingerprint achieved good results on all four metrics, but was still a little behind Xception and Capsule, especially for challenging datasets such as \textit{df\_in\_the\_wild}. Mesonet gave a similar level of performance to GAN-fingerprint on all datasets.

\begin{table*}[!h]
\centering
\vspace{-1em}
\caption{Performance ($A|P|R|F_1$) of visual forensic techniques on different datasets}
\label{tbl:accuracy_datasets}
\vspace{-5pt}
\resizebox{1.4\columnwidth}{!}{
\begin{tabular}{c|c|c|c|c|c|c|c}
    \toprule 
     & {df\_in\_the\_wild} & {Celeb-DF} & {UADFV} & {FF+}  & {DF-TIMIT} & {DFDC} & {DBD} \\
\midrule
      Mesonet & $0.65|0.68|0.57|0.62$  & $0.91|0.89|0.97|0.92$  & $0.90|0.88|0.94|0.90$ & $0.58|0.82|0.21|0.33$  & $\textbf{1.00}|\textbf{1.00}|\textbf{1.00}|\textbf{1.00}$ & $0.59|0.55|\textbf{0.96}|0.69$ & $0.94|0.92|0.91|0.91$ \\
   Capsule & $0.75|0.73|0.78|0.75$ &	$0.96|\textbf{0.96}|\textbf{0.99}|\textbf{0.97}$  &	$0.94|1.00|0.89|0.94$ &	$0.85|0.85|0.86|0.85$ &	$\textbf{1.00}|\textbf{1.00}|\textbf{1.00}|\textbf{1.00}$ &	$0.77|0.73|0.87|0.79$ & $0.96|0.96|0.95|0.97$ \\
    XceptionNet & $\textbf{0.77}|\textbf{0.76}|\textbf{0.80}|\textbf{0.77}$	& $\textbf{0.97}|\textbf{0.96}|\textbf{0.99}|\textbf{0.97}$	& $0.94|1.00|0.88|0.93$	& $\textbf{0.92}|\textbf{0.90}|0.96|\textbf{0.92}$	& $\textbf{1.00}|\textbf{1.00}|\textbf{1.00}|\textbf{1.00}$	& $\textbf{0.79}|0.74|0.91|\textbf{0.81}$ & $\textbf{0.99}|\textbf{0.99}|\textbf{0.99}|\textbf{0.99}$\\
    GAN-fp & $0.63|0.63|0.62|0.62$	& $0.86|0.85|0.99|0.91$	& $\textbf{1.00}|\textbf{1.00}|\textbf{1.00}|\textbf{1.00}$	& $0.86|0.86|0.93|0.88$	& $0.99|\textbf{1.00}|\textbf{1.00}|0.99$	& $0.67|\textbf{0.85}|0.4|0.54$ & $0.90|0.93|0.89|0.92$ \\
    FDBD & $0.63|0.60|0.77|0.67$	& $0.62|0.62|0.58|0.59$	& $0.72|0.74|0.67|0.70$ &	$0.50|0.50|\textbf{0.98}|0.66$	& $0.99|0.99|0.99|0.99$	& $0.57|0.56|0.64|0.59$ & $0.92|0.90|0.93|0.90$\\
    HPBD & $0.53|0.56|0.62|0.58$	& $0.62|0.62|0.69|0.65$	& $0.72|0.74|0.67|0.70$	& $0.55|0.50|\textbf{0.98}|0.66$	& $0.46|0.28|0.33|0.30$	& $0.5|0.49|0.20|0.28$ & $0.50|0.50|0.50|0.50$\\
    VA & $0.61|0.59|0.68|0.63$	& $0.65|0.63|0.71|0.66$	& $0.67|0.86|0.42|0.56$	& $0.59|0.58|0.62|0.59$	& $0.63|0.67|0.54|0.59$ & $0.62|0.63|0.56|0.59$	& $0.50|0.50|0.50|0.50$ \\
    \bottomrule
\end{tabular}
}
\end{table*}

Traditional techniques such as FDBD, HPBD and Visual-Artifacts do not generally perform as well as deep learning techniques. HPBD was the technique that gave the worst performance, with an accuracy for some datasets such as DF-TIMIT and DFDC of below 0.5. 
One possible reason for this lack of performance is that the background and the head pose of the actors in these datasets were arbitrary. 
FDBD performed the best of the three machine learning techniques, and even achieved accuracy on a par with deep learning techniques on several datasets such as DF-TIMIT.

\subsection{Dual-benchmarking comparison}

\begin{figure*}[!h]
    \centering
      \begin{subfigure}{0.45\linewidth}
        \centering
        \includegraphics[scale=0.26]{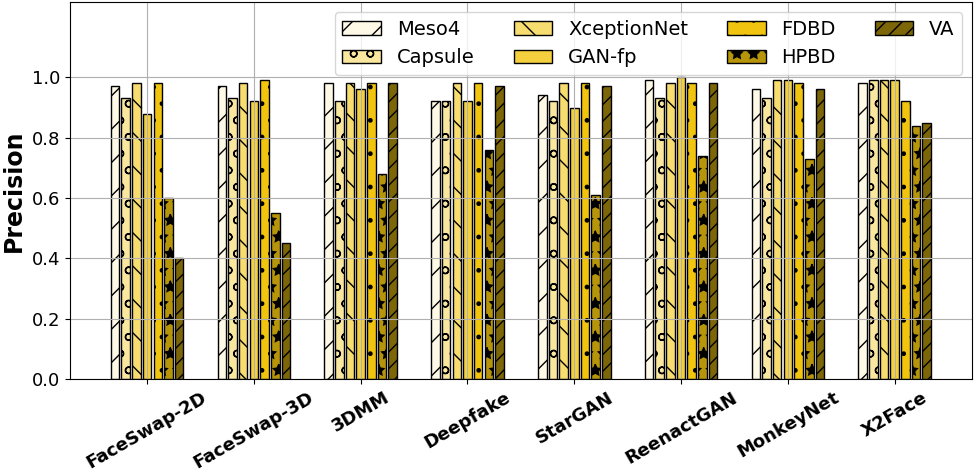}
        \caption{Precision}
        \label{fig:attr_bn}
        \end{subfigure}
        \quad \quad \quad
      \begin{subfigure}{0.45\linewidth}
        \centering
        \includegraphics[scale=0.26]{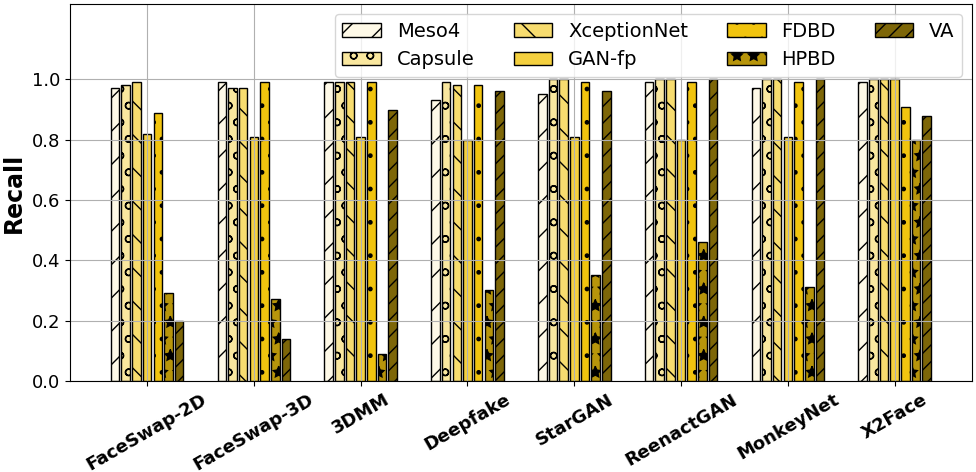}
        \caption{Recall}
        \label{fig:attr_econ}
        \end{subfigure}
    \begin{subfigure}{0.45\linewidth}
        \centering
        \includegraphics[scale=0.26]{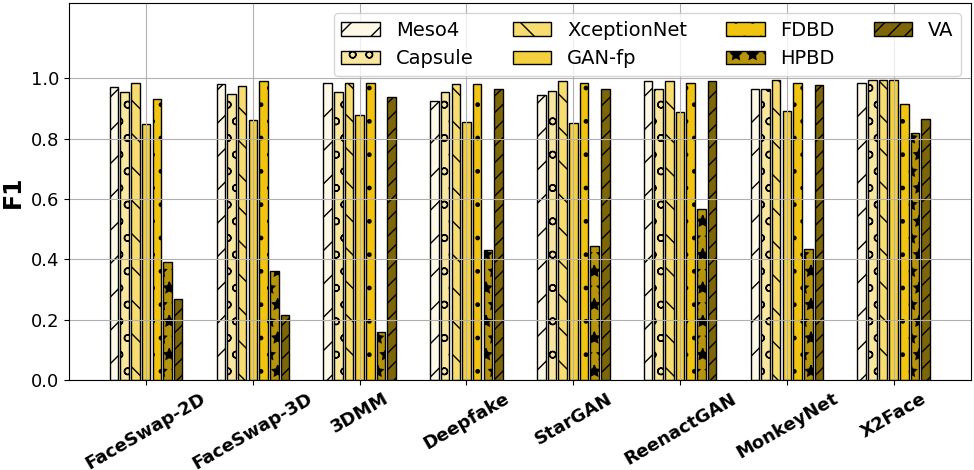}
        \caption{F1 score}
        \label{fig:attr_email}
        \end{subfigure}
     \quad \quad \quad
    \begin{subfigure}{0.45\linewidth}
        \centering
        \includegraphics[scale=0.26]{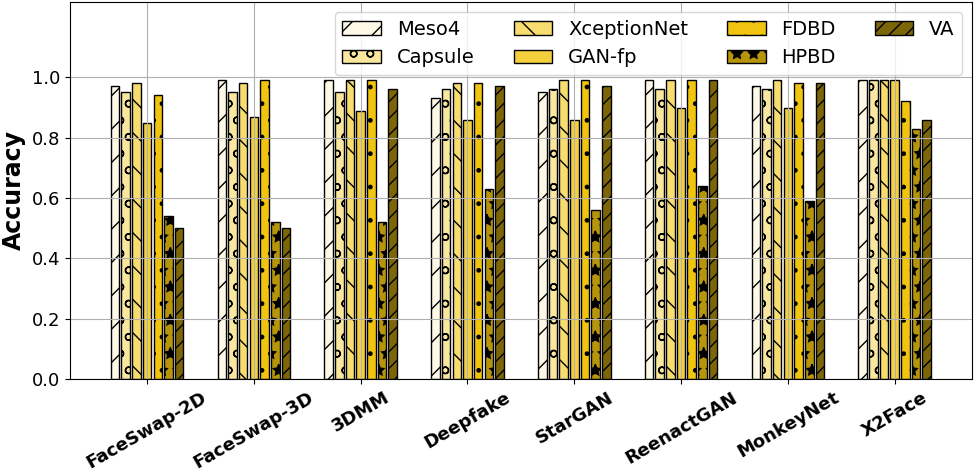}
        \caption{Accuracy}
        \label{fig:attr_email}
        \end{subfigure}
\vspace{-1em}
    \caption{Performance of visual forensics techniques against visual forgery techniques}
    \label{fig:accuracy}
 \end{figure*}

In this experiment, we analysed the fraud detection capability of forensics techniques in respect to each type of forgery. More precisely, we conducted pair-wise duels between each visual forensic technique and forgery method.
\autoref{fig:accuracy} shows the results from the use of our dual benchmarking dataset. 
It can be seen that XceptionNet performed the best overall, and that deep learning techniques generally outperformed machine learning methods. 

A more detailed examination shows that XceptionNet achieved values for precision and accuracy of greater than 90\% for all of the forgery techniques. The recall for StarGAN was the lowest of the forgery techniques, with a value of around 80\%. The performance of Capsule was fairly similar, and was slightly inferior to that of XceptionNet against each forgery technique. This is because both XceptionNet and Capsule utilise a deep neural network architecture to extract underlying patterns from the images, and the architecture of XceptionNet is more sophisticated. 
GAN-fingerprint worked surprisingly well, even for non-GAN forgeries, thanks to its in-depth analysis at the levels of both the image and model. It achieved values of 0.7-0.8 for the F1-score for all types of forgery. 

Of the machine learning approaches, it is interesting that \emph{Visual-Artifacts} countered the new generation of feature-based 
forgeries better than the traditional graphics-based forgeries. For example, Visual-Artifacts achieved F1-scores of around 0.9 on DeepFake, StarGAN and MonkeyNet, but poor F1-scores (less than 0.1) for FaceSwap-2D and FaceSwap-3D. This is because it focuses on detecting the anomalous patterns in facial details such as eyes, teeth and facial contours, which are created more accurately by the graphics-based visual forgery techniques.

\subsection{Forensic generalisation and forgery feature overlapping}

\begin{figure*}[!h]
\centering
\includegraphics[width=1.0\linewidth]{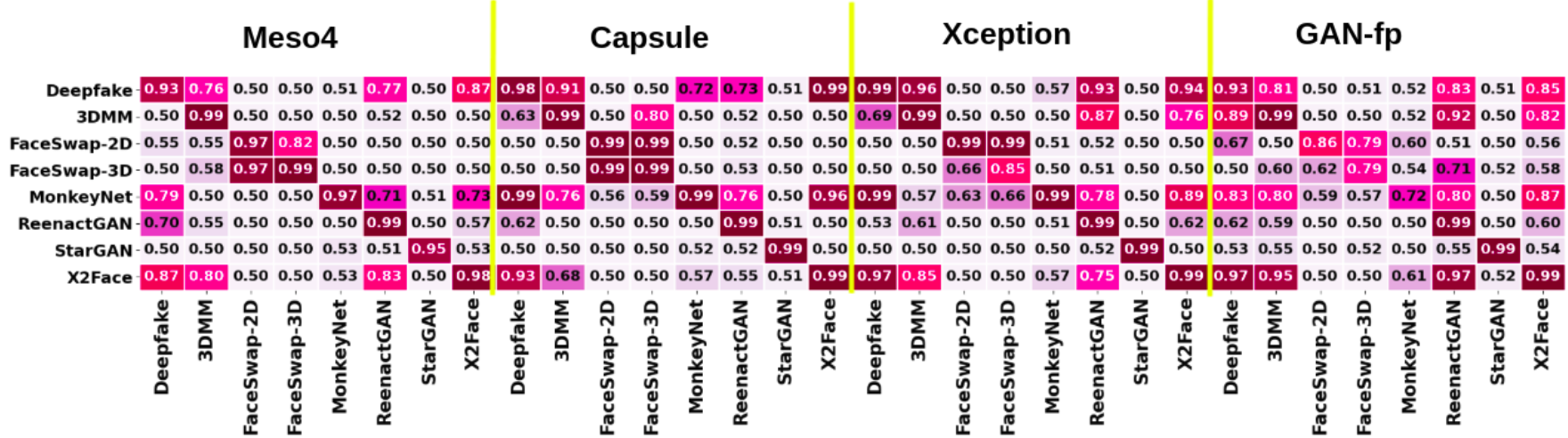}
\vspace{-10pt}
\centering
\includegraphics[width=1.0\linewidth]{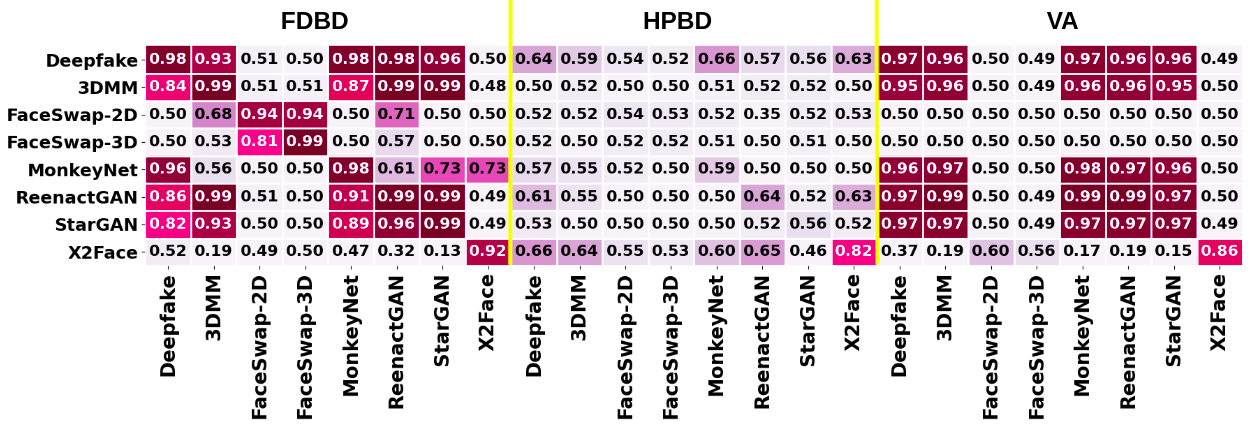}
   \vspace{-1em}
    \caption{Generalization ability of forensic techniques}
    \label{fig:generalization}
   \vspace{-1em}
\end{figure*}

In this experiment, we study the generalization ability of the forensic techniques. The \autoref{fig:generalization} shows the accuracy of the forensic models, where each forensic model was trained with one forgery type and tested on another type. In overall, forensic techniques cannot generalize well to unseen forgery, especially for traditional machine learning techniques techniques such as HPBD and FDBD, excepts Visual-Artifacts. 
This is because Visual-Artefacts leverages the expert-defined features such as global consistency (between eyes) and illumination, which gives the hint that the expert knowledge is helpful in increasing the generalization of the model. When it comes to the deep learning forensic model, Xception, GAN-fingerprint, Mesonet and Capsule gave similar performance, with Xception slightly outperformed the others. 

\begin{figure*}[!h]
    \centering
      \begin{subfigure}{0.6\linewidth}
        \centering
        \includegraphics[width=1\linewidth]{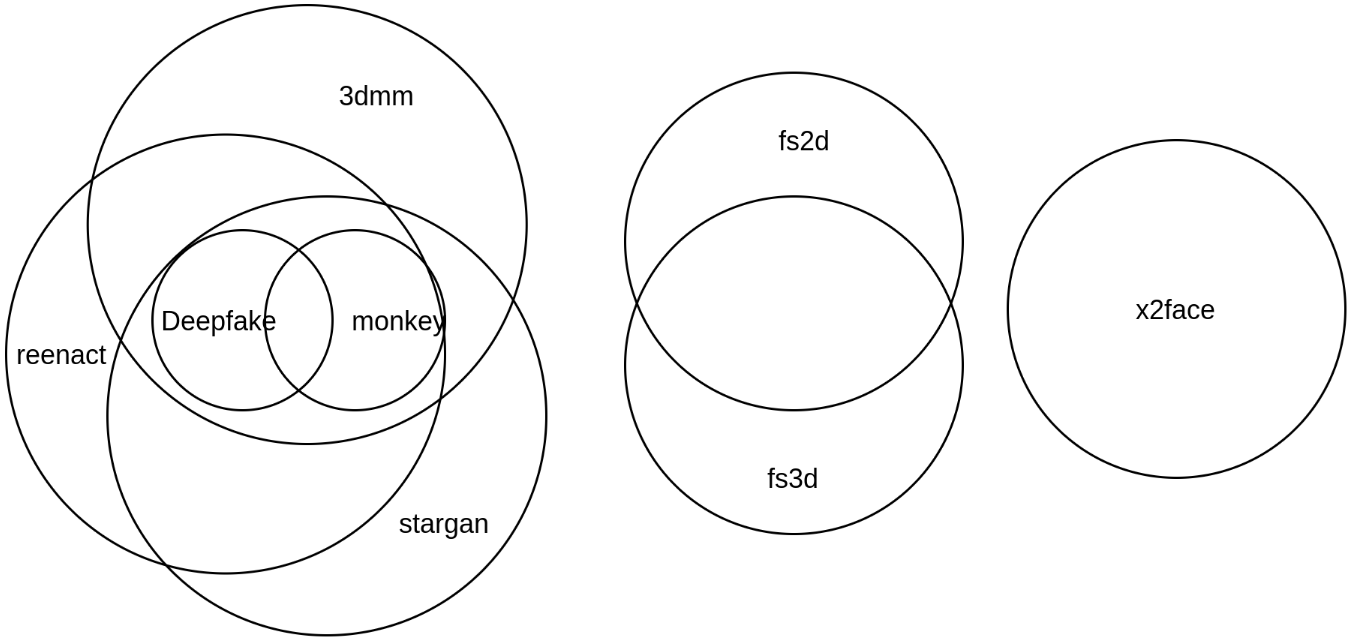}
        \caption{Under traditional forensic view}
        \label{fig:overlapping_ml}
        \end{subfigure}
        \quad \quad \quad
      \begin{subfigure}{0.6\linewidth}
        \centering
        \includegraphics[width=1\linewidth]{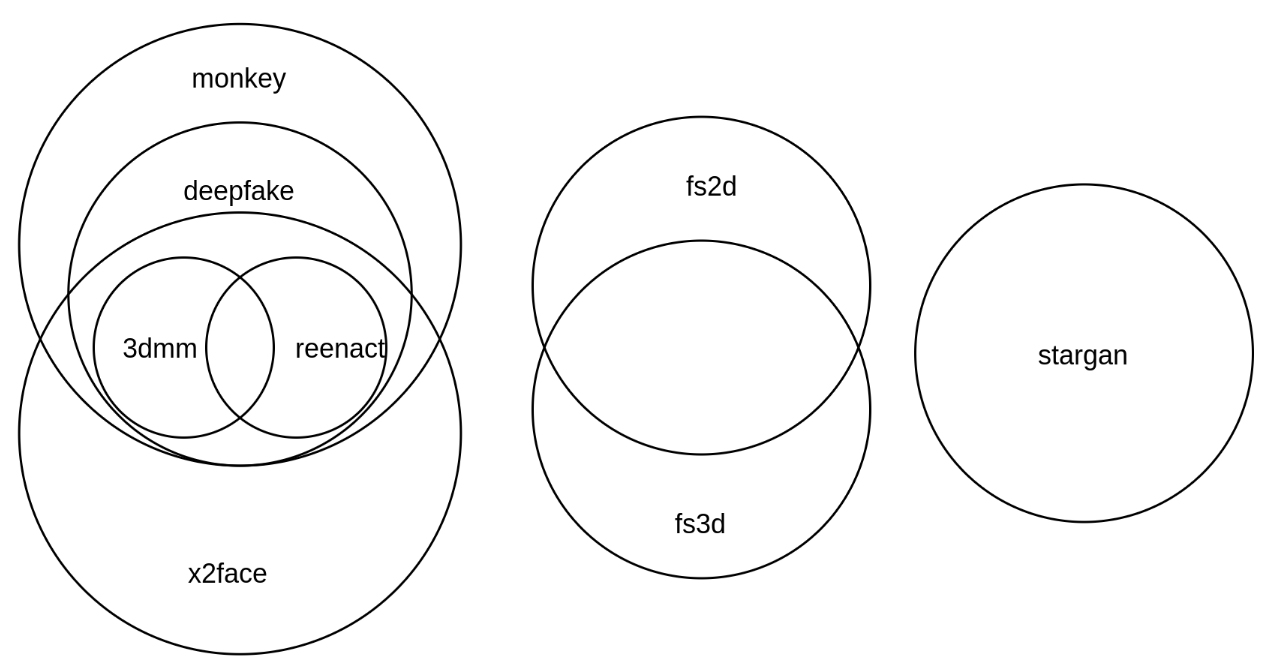}
        \caption{Under deep learning forensic view}
        \label{fig:overlapping_dl}
        \end{subfigure}
           \vspace{-1em}
    \caption{Overlapping feature of forgery techniques}
    \label{fig:overlapping_feature}
       \vspace{-1em}
 \end{figure*}

To study the overlapping forgery feature, we investigated the technique pairs that the forensic models extracted the features from one technique can perform well on the other and vice-versa, with the accuracy threshold being 0.8. The result is shown in \autoref{fig:overlapping_feature}. We realized that there was a slight difference of the forgery overlapping under the view of machine learning forensic (shown in \autoref{fig:overlapping_ml}) and deep learning forensic techniques (shown in \autoref{fig:overlapping_dl}). In more details, the graphic-based techniques FaceSwap-2D and FaceSwap-3D demonstrate the common feature in both case, as they both employ landmark-based projection. 
 However, StarGAN and X2Face demonstrated different results under machine learning view and deep learning view. This might be because the deep learning forensic techniques focus on the forgery type (attribute manipulation for StarGAN), while machine learning forensic techniques pay attention in the facial feature consistency (which is not well-maintained in X2Face).

\subsection{Qualitative study of forensic-forgery duel}

\begin{figure}[!h]
    \centering
    \vspace{-1em}
    \includegraphics[width=0.9\linewidth]{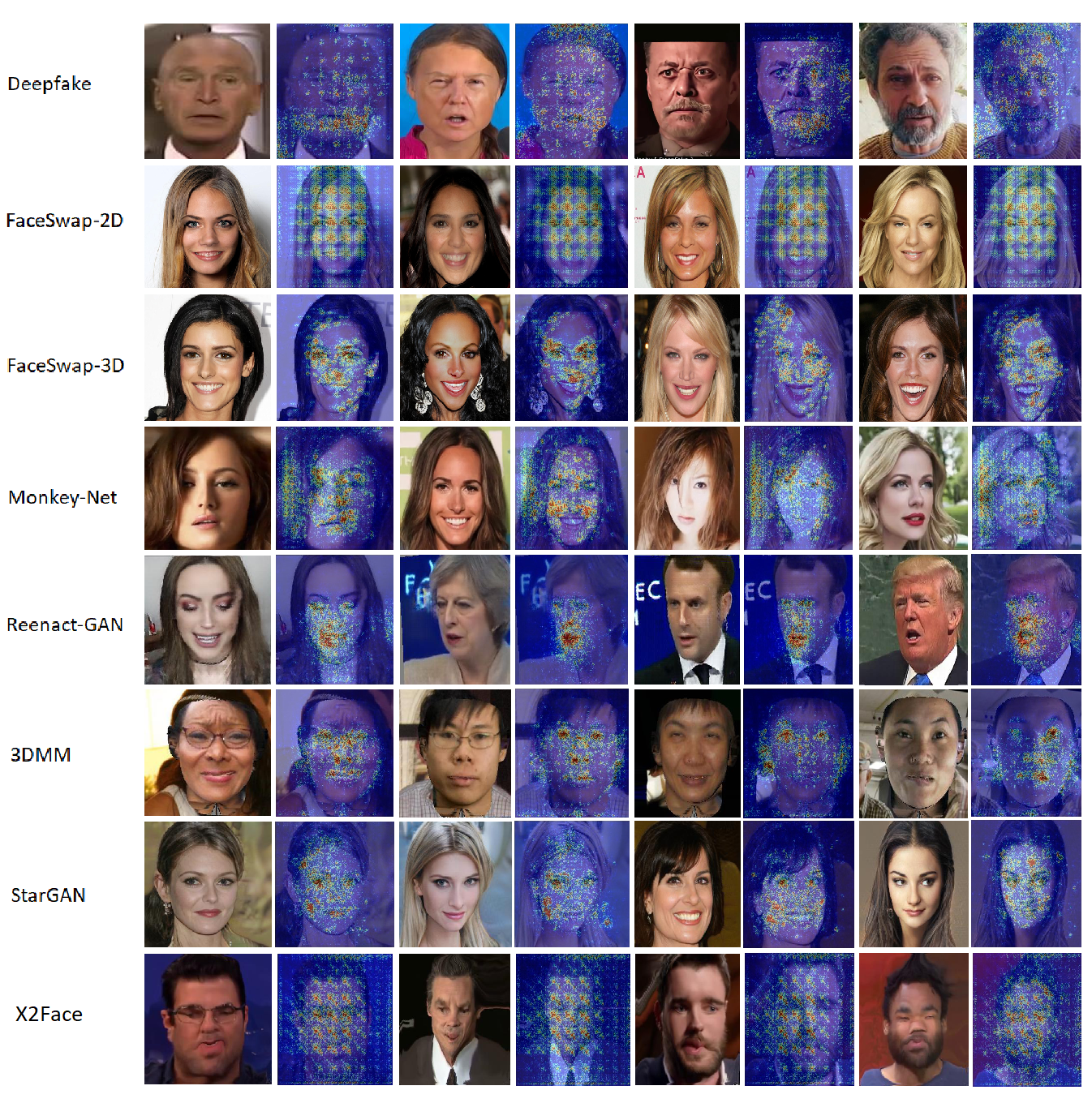}
    \vspace{-1em}
    \caption{Suspicious region of forged images}
    \label{fig:attention}
    \vspace{-1em}
\end{figure}

In this qualitative study, we learn how and where the forensic techniques detect the manipulated images. As introduced in \autoref{sec:eval_proc}, we apply an attention-based layer to the Xception-Net model to visualize the fake probability of input image pixels. The result is depicted in \autoref{fig:attention}, where we choose 4 images to demonstrate each forgery technique. It can be seen that for identity swapping techniques such as FaceSwap-2D, FaceSwap-3D and X2Face, the suspicious region were spread all the face with equal intensity. However, for DeepFake, the forgery signal is much weaker given the fact that this is an advanced GAN-based technique. On the other hand, the attention map of expression swapping techniques Monkey-Net and Reenact-GAN focus on the facial part related to the expression (mostly the mouth). For attribute manipulation techniques such as StarGAN and 3DMM, the suspicious region located mostly on the manipulated facial details such as the eyes and the cheek skin.

\subsection{Influence of contrast}

\begin{figure*}[!h]
      \begin{subfigure}{0.5\linewidth}
    \centering
      
        \includegraphics[width=\linewidth]{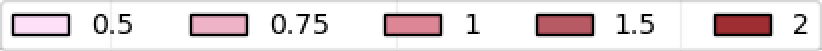} 
      
      \begin{tabular}{@{}c@{}}
        \includegraphics[width=1.0\linewidth]{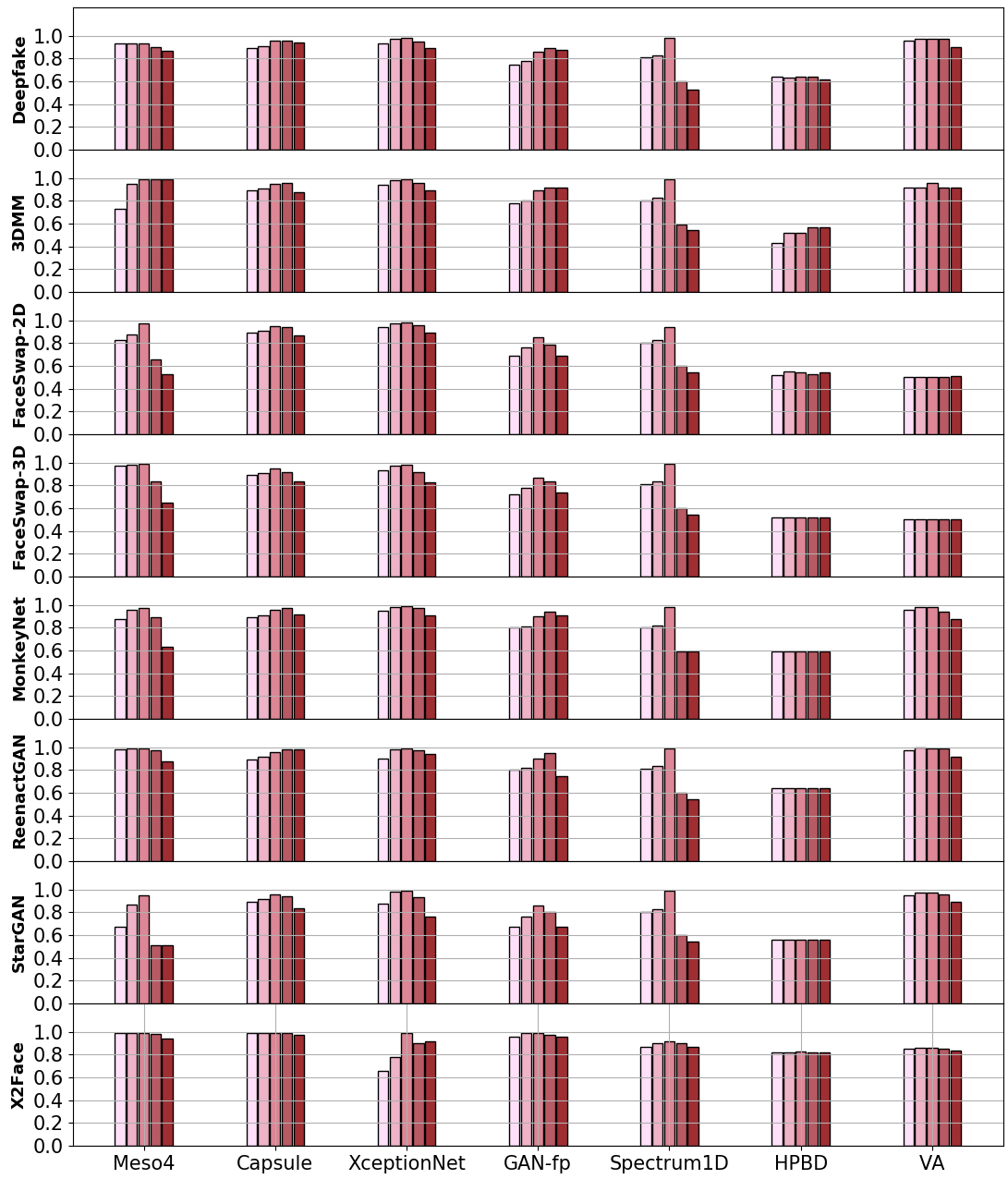} \\
      \end{tabular}
      \vspace{-10pt}
      \caption{Contrast} 
      \label{fig:contrast}
        \end{subfigure}
      \begin{subfigure}{0.5\linewidth}
    \centering
        \includegraphics[width=\linewidth]{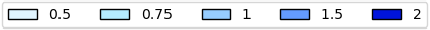} 
      
      \begin{tabular}{@{}c@{}}
        \includegraphics[width=\linewidth]{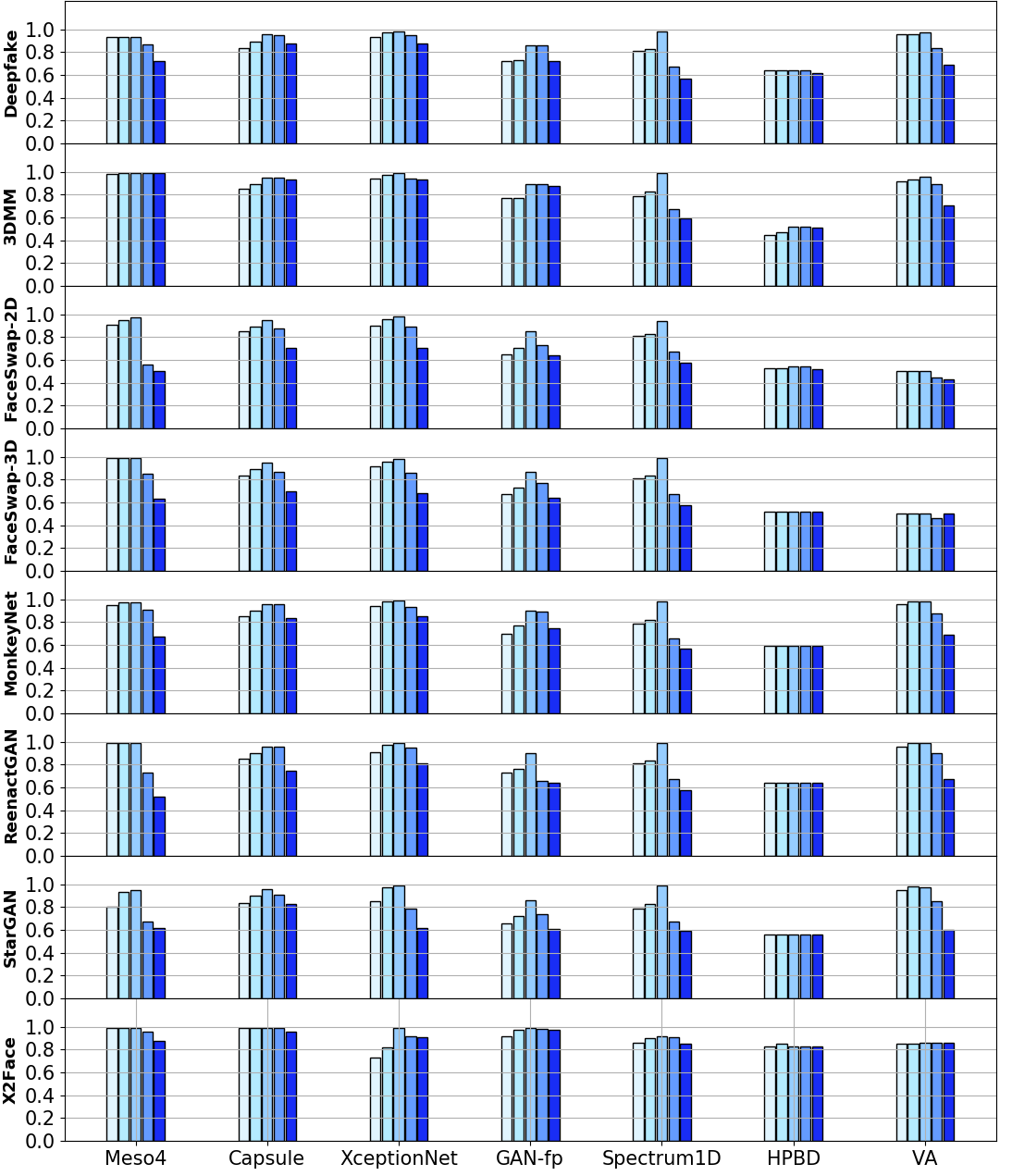} \\
      \end{tabular}
      \vspace{-10pt}
      \caption{Brightness} 
      \label{fig:brightness}
        \end{subfigure}
        \vspace{-1em}
      \caption{Effects of illumination factors} 
      \vspace{-1em}
 \end{figure*}

This experiment studied the effects of image contrast on the visual forensics techniques. \autoref{fig:contrast} illustrates the result of these experiments for a contrast factor that varied from 0.5 to 2, as described in \autoref{sec:eval_proc}. In general, all techniques suffered a reduction in accuracy when the contrast factor was at the extreme ends of this range, and performed the best when the contrast factor was 1.

XceptionNet showed the greatest robustness to this factor for all forgery techniques, and could maintain an accuracy of higher than 0.9 when the contrast factor was as high as 2 or as low as 0.5, for all forgery techniques except StarGAN. 
FDBD gave the poorest results when the contrast in the image was extreme. 
Mesonet was also unstable in terms of accuracy, due to its simple neural model architecture. The GAN-fingerprint model could in some cases achieve better accuracy when the image contrast was increased. 
This is because the higher contrast exposes more of the fingerprint of the forged image.

\subsection{Effects of brightness}

We then studied the effects of another property, the image brightness. \autoref{fig:brightness} shows the results of an experiment in which the brightness factor of the visual content was varied from 0.5 to 2. Similarly to the experiment with contrast, each technique showed a reduction in accuracy when the brightness of the visual content changed significantly.

XceptionNet showed the most technique to changes in this factor. The accuracy was stable at around 0.9 for Deepfake, 3DMM, MonkeyNet and ReenactGAN, and at around 0.7 for FaceSwap-2D and FaceSwap-3D, when the brightness factor was increased to 2. 
In contrast, Mesonet was very susceptible to changes in brightness, as its model contains significantly fewer layers than XceptionNet and Capsule. Extreme changes in brightness also adversely affected the performance of GAN-fingerprint, and its accuracy was reduced by around 0.2 for all forgery techniques when the brightness factor changed to 0.5 or 2. The traditional techniques, HPBD and Visual-Artifacts, were less susceptible to changes in the brightness of the visual content. This is because these techniques depend strongly on engineered features such as landmarks and facial details, which are unaffected by the brightness. 

\subsection{Robustness against noise}

\begin{figure*}[!h]
\centering
\includegraphics[width=1.0\linewidth]{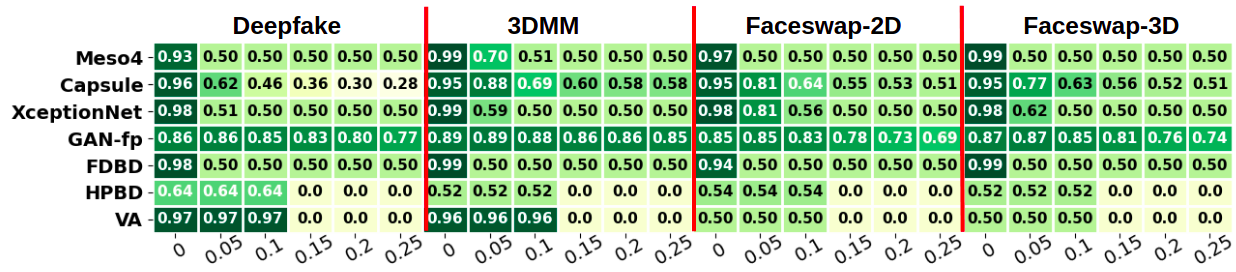}
\vspace{-10pt}
\centering
\includegraphics[width=1.0\linewidth]{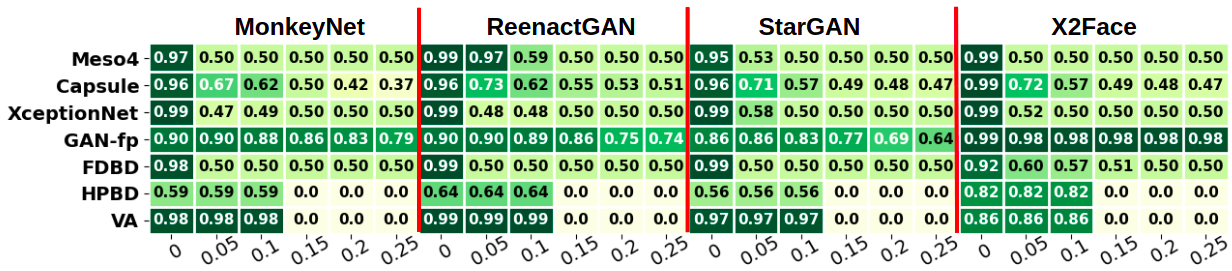}
\vspace{-1em}
    \caption{Robustness against noises}
    \label{fig:noise}
\vspace{-1em}
\end{figure*}

We then explored the effects of noise on the performance of each forensics technique. To simulate this condition, we added Gaussian noise to the images in the dual benchmarking dataset, as described in \autoref{sec:eval_proc}. \autoref{fig:noise} depicts the experimental results. An unexpected finding was that most of the forensics techniques were strongly affected by this noise factor.

GAN-fingerprint demonstrated the greatest robustness to this noise factor, and its accuracy remained above 0.7 when the noise level (standard deviation $\sigma$) reached 0.3. This can be explained by the fact that Gaussian noise does not affect the fingerprint generated by GAN-based forgery techniques, and its in-depth investigation of both the image and model level helps the model to mitigate the effects of the noise. Conversely, state-of-the-art techniques such as XceptionNet and Capsule, which showed strong potential in the previous test, did not perform well in this experiment. The accuracy of these techniques quickly fell to 0.5 when the standard deviation reached 0.1. The performance of traditional techniques such as FDBD, Visual-Artifacts and HPBD also suffered with the addition of this noise factor.

\subsection{Robustness against image resolution}

\begin{figure*}[!h]
\centering
\includegraphics[width=1.0\linewidth]{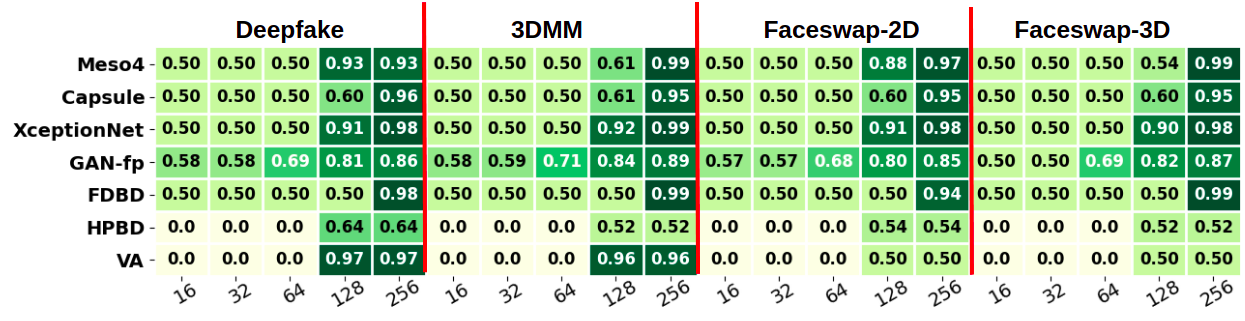}
\vspace{-10pt}
\centering
\includegraphics[width=1.0\linewidth]{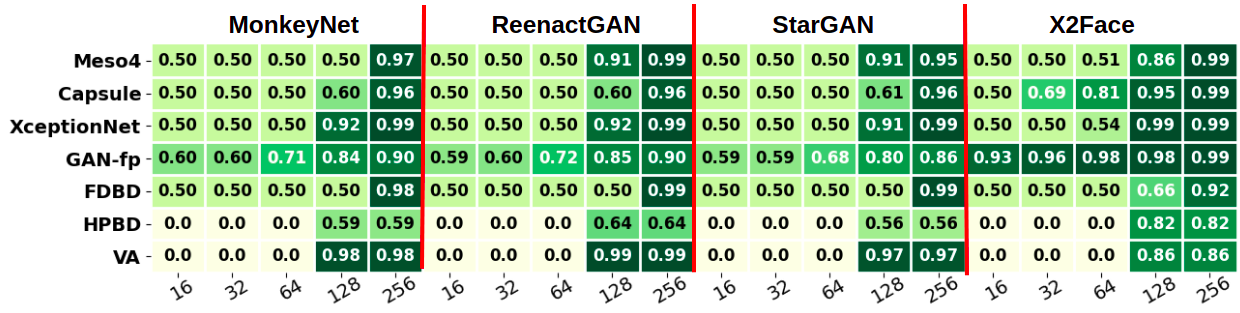}
\vspace{-1em}
    \caption{Robustness against image resolution}
    \label{fig:resolution}
\vspace{-1em}
\end{figure*}

Another important image property examined here is the resolution. \autoref{fig:resolution} shows the empirical results for each visual forensic technique when the input size of the visual content was varied from $16 \times 16$ to $256 \times 256$. It can be seen from the diagram that all visual forensics techniques perform best with high-resolution content.

A more detailed examination shows that these techniques performed better when the resolution of the input images was higher than $64 \times 64$. XceptionNet and Capsule were the two techniques that benefited the most from this increase in resolution, with an increase in accuracy from 0.5 at $64 \times 64$ to 1.0 at $256 \times 256$ for all forgery datasets. One possible reason for this is that high-resolution images allow these models to better capture the underlying latent features in the input images and video frames.

\subsection{Influence of missing information}

\begin{figure*}[!h]
\centering
\includegraphics[width=1.0\linewidth]{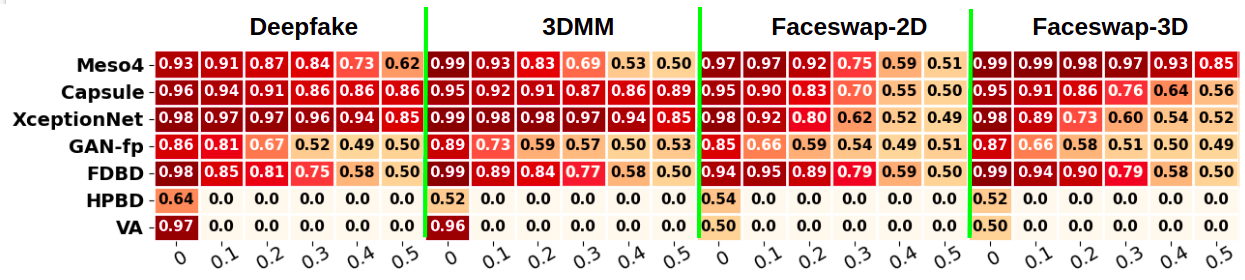}
\vspace{-10pt}
\centering
\includegraphics[width=1.0\linewidth]{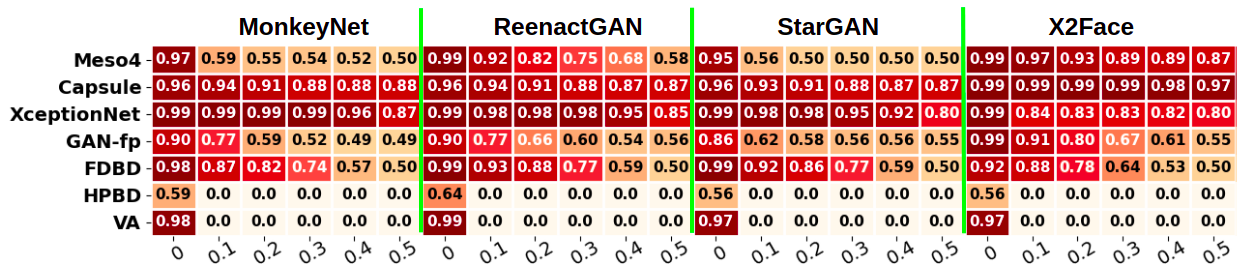}
\vspace{-1em}
    \caption{Influence of missing information}
    \label{fig:miss}
\vspace{-1em}
\end{figure*}

In this experiment, we studied the effects of missing information on the performance of visual forensics techniques. This situation was simulated using the masking strategy described in \autoref{sec:eval_proc}. \autoref{fig:miss} illustrates the experimental results when the missing pixel ratio was increased from 0.1 to 0.5. All of the forensics techniques showed a reduction in accuracy when the ratio of missing information was high.

XceptionNet and Capsule maintained a  high level of accuracy when the amount of missing information was increased, with a value of around 0.9 when the missing ratio was 0.5 for the DeepFake, 3DMM, MonkeyNet, ReenactGAN and StarGAN forgery datasets. However, for FaceSwap-2D and FaceSwap-3D, the traditional \emph{FDBD} method slightly outperformed XceptionNet and Capsule in terms of accuracy by $\approx0.05$. This is because the anomalous patterns in the frequency domain are not affected by missing information. 

\subsection{Adaptivity to image compression}

The last property investigated here is image compression. Compression algorithms are very popular, since they can significantly reduce the image size while keeping the quality at an acceptable level. In this experiment, we explored different compression settings, as described in \autoref{sec:eval_proc}.

\begin{figure}[!h]
\centering
\includegraphics[width=0.6\linewidth]{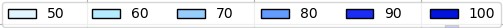}
\includegraphics[width=0.9\linewidth]{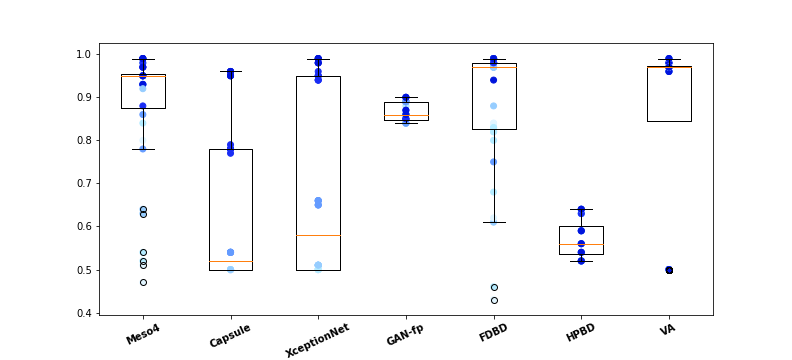}
\vspace{-1em}
    \caption{Adaptivity to image compression}
    \label{fig:compression}
\vspace{-1em}
\end{figure}

The results are shown in \autoref{fig:compression}. It can be seen that in general, the performance of all forensic techniques suffered from the loss of information produced when the compression ratio was high. Deep learning techniques such as XceptionNet, Capsule and Mesonet gave lower performance under image compression, with the accuracy falling to 0.5 when the compression ratio reached 80\%. However, GAN-fingerprint was nearly immune to this change; its accuracy remainted at a fairly high level of more than 0.8 when the compression ratio reached 50\%. Traditional techniques such as HPBD and Visual-Artifacts also showed great robustness to this factor, while the performance of FDBD underwent a considerable drop of 0.2 when the image compression ratio was high, for all forgery datasets.

\section{Conclusions}
\label{sec:con}

This paper reports the first large-scale reproducible dual benchmarking study of visual forgery and visual forensics. We analyse a range of different visual forgery and forensic paradigms and discuss the characteristics of representative models. We then design an extensible framework and construct a novel dataset that covers a large range of forgery contents. We then conduct a dual benchmarking comparison and analyse the results to provide comprehensive guidelines for researchers and users.

\sstitle{Performance guidelines}
We present a set of guidelines based on our experimental findings to help end-users find an appropriate forensics solution for a particular application requirement:
\begin{itemize}

  \item Overall, deep learning techniques such as XceptionNet and Capsule were the best. In ideal conditions, such as high resolution without noise, they can defeat even the state-of-the-art forgery algorithms such as DeepFake and StarGAN, with $\geq 90\%$ precision and accuracy. However, in cases of noise or low resolution, the effectiveness of detection was reduced. Noise and low resolution are therefore still challenges for visual forensics~\cite{qureshi2019hyperspectral}.
  
  \item Existing forensic techniques still struggle to handle unseen forgery techniques. The forensic techniques only work well when the training data contains fake images generated by the forgery technique itself or similar techniques. Expert knowledge feature can mitigate the situation, with Visual-Artifact being a typical success case.
  
  \item There exists overlapping feature between forgery techniques. This overlapping can be found when the feature extractor of forensic model for a class of fake images can be effectively used to identify another class of fakes. Forgery methods using similar core technique (e.g. GAN) often share the overlapping feature. Besides, the forgery overlapping effectiveness also depends on the property of the forensic method. 
  
  \item The forensic of forged images can be visualized and analyzed using tool such as attention map. For different forgery type, forensic techniques have different behaviour when determine the suspicious zone of the forged images. For example, the suspicious zone is often spread in the whole face for identity swapping, while for expression swapping the suspicious zone is focused on the certain facial part such as mouth and eye. 
  
  \item For low-quality or low-resolution input data, we recommend the GAN-fingerprint forensics algorithm, as it is robust against extreme input conditions. This technique can maintain an accuracy of above 0.6 when the content resolution falls to just $16 \times 16$ and the added noise level (standard deviation) is 0.3. However, to guarantee standard performance from visual forensic techniques, we recommend that the input resolution should be at least $128 \times 128$.
  
  \item In terms of the processing speed, XceptionNet and Capsule were the fastest, requiring only 23 and 26s, respectively, to process 10K images. XceptionNet was slightly better in overall, but Capsule is a significant less complex model with six times fewer parameters. Mesonet also has a fast processing speed, but the trade-off between speed and prediction quality is significant. Based on our empirical results, we recommend XceptionNet and Capsule in preference to Mesonet. 
  
  \item Machine learning techniques such as FDBD, HPBD and Visual-Artifacts do not involve sophisticated models or training processes, but their prediction results are less reliable, and are more susceptible to adverse factors. These forensics techniques are also weak against graphics-based forgery techniques such as FaceSwap-3D, which focuses on maintaining the local facial characteristics. To maintain quality, we do not recommend using ML techniques. 
  
  \item Extreme brightness and contrast of the input content can adversely affect the performance of visual forensics techniques. If the levels of these two factors are significant, the user should consider XceptionNet as it is least susceptible to these factors. 
\end{itemize}

These findings are summarised in \autoref{tbl:guideline}, where the best, the second-best and the worst techniques are shown for each performance category.

\begin{table}[H]
\centering
\footnotesize
\vspace{-1em}
\caption{Performance guideline for visual forensics}
\vspace{-1em}
\label{tbl:guideline}
\resizebox{0.6\columnwidth}{!}{
\begin{tabular}{lccc}
    \toprule 
     \textbf{category} & \textbf{winner} & \textbf{1st runner-up} & \textbf{worst} \\
    \midrule
    accuracy &  XceptionNet  &  Capsule  & HPBD \\
    precision &  XceptionNet  &  Capsule & HPBD \\
    recall &  FDBD  &  Capsule & HPBD \\
    contrast &  XceptionNet  & Capsule  & HPBD \\
    brightness &  XceptionNet  & Capsule  & HPBD \\
    noises & GAN-fingerprint   &  Visual-Artifacts & VA \\
    resolution & GAN-fingerprint   & XceptionNet  & VA \\
    missing information & XceptionNet   & Capsule  & VA \\
    compression &  FDBD  & GAN-fingerprint  & VA \\
 \bottomrule
 \end{tabular}
 }
\end{table}

\begin{figure}[!h]
    \centering
    \includegraphics[width=0.8\linewidth]{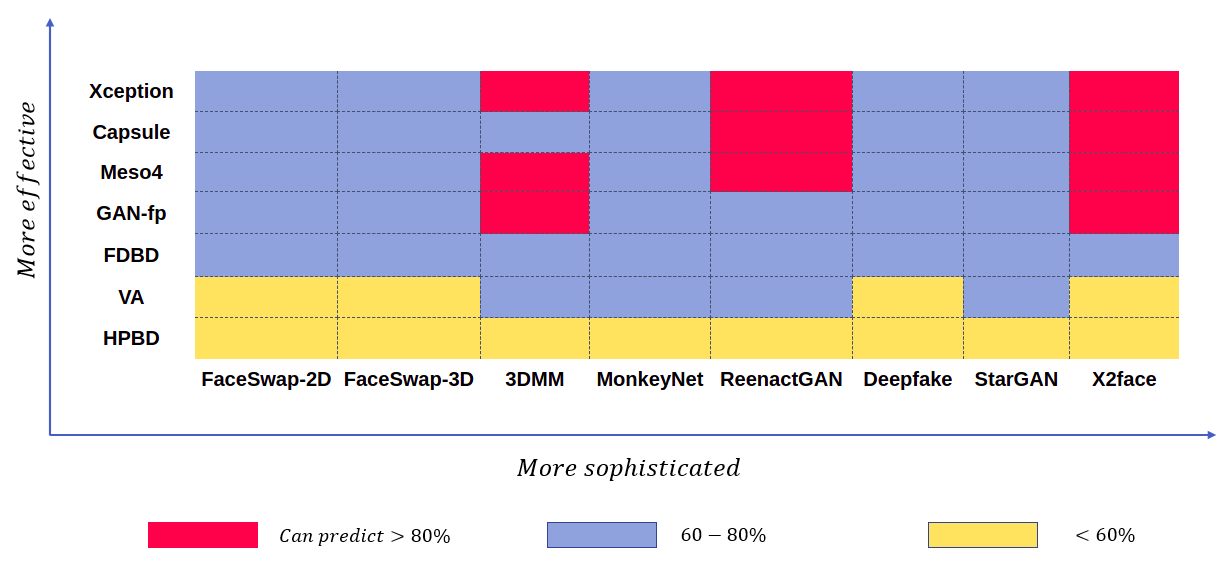}
    \caption{Dual benchmarking guideline on forensic and forgery}
    \label{fig:dual}
\end{figure}

\sstitle{Dual-benchmarking guidelines}
These results are illustrated in \autoref{fig:dual}, in which the performance of more effective techniques such as XceptionNet, Mesonet and Capsule are highlighted, as they achieved high accuracy for all forgery techniques. On the other hand, techniques such as  HPBD showed less impressive result, and their selection for real-world applications should therefore be considered carefully. 

\begin{figure}[!h]
    \centering
    \includegraphics[width=0.8\linewidth]{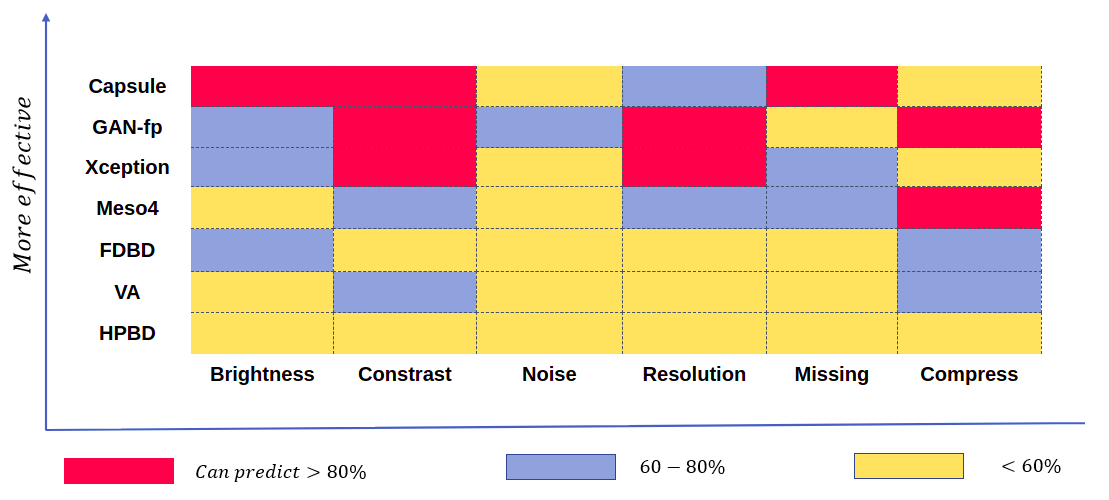}
    \caption{Robustness of visual forensic techniques against extreme adversarial factors}
    \label{fig:adversarial_factor}
\end{figure}

In terms of adversarial conditions (brightness, contrast, noise, resolution, missing information, compression), Capsule and GAN-fp are the most robust techniques, while Xception and Meso4 have moderate performance. Hand-crafted models such as FDBD, VA and HPBD degrade significantly in these settings. The summary of visual forensic techniques' robustness against adversarial conditions is shown in \autoref{fig:adversarial_factor}.

\sstitle{Future work} Since the war between visual forgery and visual forensics is never-ending, our framework can act as a tool for monitoring, comparing, and selecting the most appropriate countermeasure for a particular scenario. In future work, we expect that our open-source benchmark will be enriched by the research community and decision makers with new datasets and models. Also, more thorough investigation about the video-specific forged content detection would be an interesting consideration.


\end{document}